\documentclass[]{interact}

\usepackage{epstopdf}% To incorporate .eps illustrations using PDFLaTeX, etc.
\usepackage[natbibapa,nodoi]{apacite}% Citation support using apacite.sty. Commands using natbib.sty MUST be deactivated first!
\theoremstyle{plain}% Theorem-like structures provided by amsthm.sty

\theoremstyle{definition}

\theoremstyle{remark}

\usepackage{graphicx}
\usepackage[figurename=Figure,labelfont=bf,labelsep=period]{caption}
\usepackage{subcaption}
\usepackage{amsmath}
\usepackage{algorithm}
\usepackage{algpseudocode}
\usepackage{multirow}
\usepackage{booktabs}
\usepackage{array}
\usepackage{longtable}
\usepackage{pdflscape}
\usepackage{ragged2e}
\usepackage{adjustbox} 
\usepackage{tikz}
\usepackage{dsfont}
\usepackage[colorlinks=true, citecolor=black, linkcolor=black, urlcolor=black]{hyperref}

\usepackage{setspace}
\begin{document}

\title{Hierarchical Deep Reinforcement Learning Framework for Multi-Year Asset Management Under Budget Constraints\thanks{This is a preprint of a manuscript submitted to Structure and Infrastructure Engineering (Taylor \& Francis) and is currently under review.}}

\author{
\name{Amir Fard\textsuperscript{a} and Arnold X.-X. Yuan\textsuperscript{a}\thanks{CONTACT Arnold Yuan. Email: arnold.yuan@torontomu.ca}}
\affil{\textsuperscript{a}Department of Civil Engineering, Toronto Metropolitan University \\
\,\,350 Victoria Street, Toronto, ON, Canada, M5B 2K3}
}

\maketitle

\begin{abstract}

Budget planning and maintenance optimization are crucial for infrastructure asset management, ensuring cost-effectiveness and sustainability. However, the complexity arising from combinatorial action spaces, diverse asset deterioration, stringent budget constraints, and environmental uncertainty significantly limits existing methods' scalability. This paper proposes a Hierarchical Deep Reinforcement Learning methodology specifically tailored to multi-year infrastructure planning. Our approach decomposes the problem into two hierarchical levels: a high-level Budget Planner allocating annual budgets within explicit feasibility bounds, and a low-level Maintenance Planner prioritizing assets within the allocated budget. By structurally separating macro-budget decisions from asset-level prioritization and integrating linear programming projection within a hierarchical Soft Actor-Critic framework, the method efficiently addresses exponential growth in the action space and ensures rigorous budget compliance. A case study evaluating sewer networks of varying sizes (10, 15, and 20 sewersheds) illustrates the effectiveness of the proposed approach. Compared to conventional Deep Q-Learning and enhanced genetic algorithms, our methodology converges more rapidly, scales effectively, and consistently delivers near-optimal solutions even as network size grows. 

\end{abstract}


\newpage 

\doublespacing

\section{Introduction}

Effectively and efficiently managing infrastructure networks is crucial for sustainable development, economic progress, and public safety. Assets such as transportation systems, water and wastewater networks, and energy grids are integral to modern societies, supporting daily activities and enabling large-scale services. However, a variety of forces drive the urgency of proper infrastructure asset management, including natural deterioration processes, the increasing impact of climate change, technological advancements, and growing demands that stem from urbanization and economic expansion. These pressures underscore the need for comprehensive and adaptive strategies to maintain and improve infrastructure performance over multiple years, while consistently meeting budgetary and operational constraints \citep{Arcieri2024,Hamida2024,Yao2024,lei2025integration}.

Infrastructure Asset Management (IAM) encompasses several tasks: inventorying and evaluating the condition of assets, projecting their future states, and  budgeting and prioritization \citep{yuan2017principles}. Among these tasks, budgeting and prioritization (often referred to as maintenance planning) represent a pivotal challenge, as they determine which actions, such as maintenance, rehabilitation, or replacement, should be taken for each asset and when those actions should be executed. Such decisions affect the performance of individual assets, the total costs incurred over their lifecycles, and the reliability of the entire network. Developing an optimal multi-year maintenance plan that complies within annual budget constraints becomes particularly difficult, because infrastructure networks often contain numerous interlinked assets, each with its own deterioration behaviors, and exposures to varying operating and environmental conditions \citep{Yao2024,Zhou2022_2}. 

The sequential nature of these decisions further complicates the planning task \citep{Fard2024}. Maintenance policies must account for how interventions at one point in time influence asset conditions in future periods, thereby affecting subsequent budgets and the cost-effectiveness of later actions. Additionally, non-linear relationships may arise when certain improvements or rehabilitations yield synergistic effects on the network, or when deterioration accelerates non-linearly once an asset’s condition drops below a threshold. Handling these sequential changes and non-linear behaviors often goes beyond the capacity of many classical optimization techniques.

Over past decades, problem formulation of the multi-year network maintenance, rehabilitation and reconstruction (MRR) planning has evolved from a two-stage decision-making that separates the `program-level' budget planning from the `project-level' asset prioritization \citep{Golabi1982,GaoZhang12}, to an integrated, asset-based, bottom-up framework \citep{Sathaye2012,LeeJW15JointOpt}. Although linear programming can handle a large-size asset inventory or network, it does not fit the bottom-up framework, which by nature has turned the problem to a nonlinear integer or mixed-integer program.   A range of solutions has been explored, including traditional integer programming, heuristic optimization, evolutionary algorithms, and dynamic programming. While these methods have yielded insights into particular use cases, they often fall short in addressing large, complex networks with high-dimensional state and action space. Branch and bound or cut methods can only deal with small problems; heuristic optimizations may not systematically converge to near-optimal solutions; and dynamic programming methods often confront intractable state-action spaces in large-scale infrastructure contexts.

Reinforcement Learning (RL) has emerged as a promising alternative for the network MRR planning. By allowing an agent to learn from interactions with an environment, RL approaches naturally handle the uncertainty and time-coupled decisions critical to infrastructure planning. Recent developments in Deep Reinforcement Learning (DRL) have combined RL with deep neural networks, enabling the handling of very large or continuous state-action spaces \citep{sutton2018reinforcement,Mnih2015}. These DRL techniques have proven especially valuable for problems once deemed computationally intractable because of their capacity to approximate complex value or policy functions. Within the realm of infrastructure asset management, DRL can help an agent learn cost-effective interventions directly from simulations of asset deterioration, climate variability, and other uncertain factors \citep{Lei2022,Du2022,Leppinen2025,Do2024,Sasai2024}. However, straightforward DRL implementations can still struggle when the combinatorial action spaces (for instance, binary maintain/do-nothing decisions over many assets) meet budget constraints, leading to slow convergence or infeasibility. As our review in the next section reveals, although there have already been quite a few research groups devoting to DRL algorithms for network MRR planning problems, handling budget constraints explicitly and effectively in DRL remains to be an area that has received little attention. 

Overall, the literature has progressively shifted from deterministic linear models to MDPs, POMDPs, and RL-based approaches that can better handle sequential decision-making requirements inherent in infrastructure management. Metaheuristics continue to thrive, especially when integrated with local search or linear subproblems \citep{Fard2024}, but they typically face scalability challenges when applied to extensive networks under tight multi-period budget constraints. Although RL and deep learning approaches offer significant advantages in managing high-dimensional systems, they frequently either neglect explicit budget enforcement or address it indirectly through cost-minimizing reward structures. Emerging hierarchical RL frameworks aim to address these gaps by decomposing complex decision-making processes into multiple interconnected layers. 
% One layer, typically at a higher level, manages annual budget allocations, while lower-level layers allocate maintenance interventions across individual assets within those budgets.
Such hierarchical decomposition allows for efficient handling of system-wide constraints alongside detailed asset-specific considerations, reducing complexity and enhancing interpretability of policies.

% In short, there is consensus that deep reinforcement learning, particularly in hierarchical or factorized form, is capable of addressing the complexity and scale of large, interdependent infrastructure networks. Traditional MDP-based frameworks have laid the groundwork for sequential decision-making under uncertainty, but they struggle computationally once partial observability, large state spaces, or multi-year budgets are introduced. Metaheuristics have proven versatile but, in many cases, lack the theoretical guarantees or face scaling difficulties of their own. Reinforcement learning exploits function approximation to navigate massive state and action spaces adaptively. However, ensuring interpretability, stability of training, and explicit resource limitations continues to be a challenge.

This paper proposes a Hierarchical Deep Reinforcement Learning (HDRL) framework to manage multi-year budgets and asset maintenance decisions. Built upon an understanding of the innate hierarchical structure of the network maintenance planning problem, the proposed HDRL methodology comprises two interconnected layers: a \emph{Budget Planner} at the upper tier responsible for annual budget determination, and a \emph{Maintenance Planner} at the lower tier allocating resources based on asset conditions and maintenance priorities. A local optimization step ensures selected maintenance actions are moving toward maximizing the performance of the system while the costs remain within allocated budgets. This hierarchical decomposition significantly reduces the exponential growth of combinatorial actions associated with single-agent RL approaches while effectively capturing sequential dependencies and nonlinear asset behaviors.

To validate the proposed HDRL framework, we conduct a comparative analysis through a sewer network case study with varying complexities involving 10, 15, and 20 sewersheds. We benchmark HDRL performance against constraint programming for the smaller instance and evaluate scalability relative to a Deep Q-Learning (DQL) baseline for larger instances. Results demonstrate HDRL's robust convergence, superior stability, and scalability advantages compared to conventional RL methods.

The remainder of this paper is structured as follows: Section 2 formulates the problem of multi-year maintenance planning under budget constraints and covers the related works. Section 3 provides a comprehensive overview of RL and DRL, emphasizing relevant methodologies such as policy gradients and Soft Actor-Critic (SAC). Section 4 details the proposed hierarchical DRL approach, including its architecture and training procedures. Section 5 presents the case study results, followed by discussions on scalability and implications. Finally, Section 6 concludes the paper and suggests future research directions.

\section{Review of Existing Methodologies}
Budget planning and maintenance prioritization is a hard research problem that has been studied by several generations. A good analysis of the problem and traditional methods related to pavement management can be found in \citet{GaoZhang12}. The primary purpose of this section is to present a brief review of the major work related to the applications of deep reinforcement learning to multi-year budget planning and network maintenance prioritization in the context of infrastructure asset management. Before this, a canonical model of the problem is presented below. 

\subsection{Problem Setting and Canonical Model}

Infrastructure Asset Management (IAM) involves planning multi-year interventions, such as maintenance, rehabilitation, and replacement, to keep a network of assets in desired condition while respecting budgetary constraints. Consider a planning horizon of \( h \) years and \( n \) infrastructure assets, where each asset \( i \) (\( i = 1,\dots,n \)) has a performance measure \( s_{i,t} \) at the beginning of year \( t \). Note that the asset-level performance measure can be multivariate and in this case \( s_{i,t} \) would be a vector. In this paper, we assume it univariate and treat it as the state variable of the asset. 

Each asset’s state evolves over time according to whether a maintenance action is performed. The binary decision variable \(x_{i,t}\in\{0,1\}\) indicates if a maintenance action is applied to asset \(i\) in year \(t\). Asset \(i\) then transitions from \(s_{i,t}\) to \(s_{i,t+1}\) via the following transition relationship, also known as a deterioration model:
\begin{equation}
\label{eq:state_transition}
s_{i,t+1} 
= 
x_{i,t}\,f_{i}^{\text{act}}\bigl(s_{i,t}\bigr)
+ 
\bigl(1 - x_{i,t}\bigr)\,f_{i}^{\text{det}}\bigl(s_{i,t}\bigr),
\end{equation}
where \(f_{i}^{\text{act}}\) describes the effect of applying intervention, and \(f_{i}^{\text{det}}\) characterizes how the asset naturally deteriorates if left untreated. Note that a deterioration model often consists of two parts: a natural deterioration part and a maintenance effectiveness part. 

Due to budgetary and other constraints, the applications of asset-level maintenance activities must be coordinated. To facilitate the coordination, a network-level performance measure, often known as Level of Service (LoS), needs to be defined based on the state of every asset. 
%performance function \( \kappa(\cdot) \) that maps the asset’s condition to a scalar index, which can represent, for example, pavement ride quality based on a relevant standard or a bridge’s structural integrity. Each asset \( i \) is also assigned a weight \( w_i \) (e.g., length or area) to reflect its relative importance within the network. The principal goal is to maximize the expected total performance of all assets over the \( h \)-year horizon. Define the network's level of service (LoS) at time $t$ as
A general expression for LoS at time $t$ is given as:
\begin{equation}
\mathrm{LoS}_{t}
=\,
\sum_{i=1}^{n} w_{i}\,\mathbb{E} \, \bigl[ \kappa \bigl(s_{i,t} \bigr) \bigr],
\end{equation}
where \(w_{i}\) is the relative weight (length, area, or importance) of asset \(i\), \(\kappa(\cdot)\) is a function that coverts the asset-level condition to a univariate contribution to the system performance. For example, when the performance function is an identity function, i.e., \(\kappa(s_{i,t}) = s_{i,t}\), the LoS represents the inventory's average performance. If \(\kappa(s_{i,t}) =  \mathds{1}_{\{s_{i,t} \, \le \, \xi_p\}}\), where $\xi_p$ is the condition threshold for a poor asset, the LoS represents the inventory's percentage of poor asset at year $t$. This generalized LoS definition can also adapt to multivariate asset performance $s_{i,t}$. 

 The principal goal of network maintenance planning is to maximize the expected total or average performance of all assets over the whole planning horizon. In addition, there are budgetary constraints. Let \(c_{i,t}\) be the unit cost of maintaining asset \(i\) in year \(t\). If maintenance is performed on asset \(i\), the total cost is \(c_{i,t}\,w_i\). Expenditures in year \(t\) must remain between a lower limit \(b_{t}^{l}\) and an upper limit \(b_{t}^{u}\), and the cumulative expenditure cannot exceed \(b^{\text{total}}\). Therefore, the multi-year IAM planning can be formulated as the following optimization problem: 
\begin{eqnarray}
    \max & & \frac{1}{h} \sum_{t=1}^{h} \mathrm{LoS}_{t} \label{eq:objective} \\  
   \mathrm{subject \,\, to} &&   \nonumber \\
  & & \sum_{i=1}^{n} c_{i,t} \, w_i \, x_{i,t} \;\geq\; b^l_t, \quad \forall\,t = 1, \dots, h, \nonumber \\
  & &  \sum_{i=1}^{n} c_{i,t} \, w_i \, x_{i,t} \;\leq\; b^u_t, \quad \forall\,t = 1, \dots, h, \nonumber \\ 
 &&  \sum_{t=1}^{h} \sum_{i=1}^{n} c_{i,t} \, w_i \, x_{i,t} \;\leq\; b^{\text{total}}. \nonumber
\end{eqnarray}

The above binary treatment  can be readily generalized to include multiple interventions per asset either by expanding the value range of $x_{i,t}$ to a general integer or by introducing an additional `method' to make  \(x_{i,j,t}\) for each asset \(i\), intervention type \(j\), and year \(t\). In the latter case, an additional assignment constraint,
\( \sum_{j \in \,\mathcal{J}_{i}} x_{i,j,t} = 1 \, (\forall i,t), \)
enforces that exactly one intervention (including ``do nothing'' as an option) is selected for each asset~\(i\) in each year~\(t\). 

The use of binary variables introduces a combinatorial aspect, especially when assets are managed over multiple years. Decisions made in earlier years directly affect the conditions of assets in subsequent years, creating intertemporal dependencies; see Equation (\ref{eq:state_transition}). In addition, the functions \(f_{i}^{\text{act}}\) and \(f_{i}^{\text{det}}\) can be nonlinear, and the deterioration process may be stochastic, making the objective’s expectation term potentially difficult to handle. Overall, the mathematical prototype presented above is an integer nonlinear optimization problem. Due to the nonconvexity, the problem is notoriously hard to solve, particularly when the size of the infrastructure network gets large.

% \subsection{Traditional Methods}
% For multi-year pavement rehabilitation planning, \citet{ouyang2004optimal} tried the problem using two solution methods, a custom branch-and-bound algorithm and a greedy heuristic. The exact branch-and-bound approach could find optimal solutions for a small case, but it was computationally expensive, motivating the heuristic method. Notably, the heuristic produced near-optimal results at much lower computational cost. In fact, their results showed that the greedy heuristic sometimes provided a better solution than the branch-and-bound search within a comparable time duration, underscoring the difficulty of solving the nonlinear model to optimality. This outcome illustrates that, while the traditional mathematical programming formulation precisely describes the problem, general-purpose nonlinear solvers struggle as the problem size grows. 

To alleviate the computational burden of nonlinear optimization, many researchers tried to reformulate it as an integer or mixed-integer program. Tractability is usually obtained by imposing simplifying assumptions, such as treating the entire network as homogeneous or discretising the condition state \citep{kuhn2005model,Golabi1982,wu2009pavement,gomes2022integer}. For instance, \citet{kuhn2005model} assume that all facilities share identical deterioration behaviour, size, and cost profiles. Most of these formulations pursue a system-level perspective, and the resulting maintenance strategies are referred as randomized policies, a concept first introduced by \citet{Golabi1982}. Another common simplification is to limit each asset to at most one intervention throughout the planning horizon \citep{ezzati2021integrated}. Although such assumptions substantially reduce computational effort, they may diverge from the heterogeneity and complexities encountered in real-world asset management practice.

Because of this, a vast body of literature in infrastructure asset management has resorted to heuristic and metaheuristic approaches instead  \citep{fwa1996genetic,chan2001constraint,Santos17,allah2020multi,altarabsheh2023hybrid}. Although these approaches  can often find near-optimal or at least feasible and satisfactory solutions within practical runtimes, they do not guarantee an optimal solution. Most recently, \citet{Fard2024} proposed a hybrid approach by combining local LP solutions with global heuristic search. 

The canonical formulation of the network maintenance optimization has an important Markov decision process feature, making it particularly suitable for reinforcement learning frameworks. Note that this formulation assumes perfect observations and completely certain deterioration models. When measurement errors or model uncertainties need or both need to be considered, the problem turns to a partially observable Markov decision process (POMDP). Even so, it can still be formulated into a belief-based MDP; for more details, see \citet{Andriotis2021} and \citet{Arcieri2024}. As modern deep reinforcement learning techniques emerged, characterized by the development of neural network, the infrastructure asset management community has over recent few years witnessed a fast growth in the use of various deep reinforcement learning methods to solve the network maintenance optimization. It is not our intention to provide here a comprehensive review, for which interested readers can refer to \citet{Asghari22} and \citet{Marugan23}. In the following, our review focuses on two aspects of the development, namely scalability and constraints handling, as they are the most critical factor to the success of an DRL method for budget-constrained network MRR planning of large-scale asset inventories.  

\subsection{Decomposition for scalability}
\emph{Divide and Conquer} is an enduring problem-solving wisdom and algorithm principle. Different decomposition approaches yields different scalability. Three categories of decomposition techniques have been used in DRL for network MRR planning. 

\subsubsection{Temporal decomposition via sequential decision making}
Casting multi-year IAM planning as a Markov Decision Process is itself a decomposition approach: the horizon-wide mixed-integer optimization is split into a sequence of annual sub-problems.  At each stage the decision-maker decides an action vector \(x_{1:n,t}\), observes stochastic state transitions governed by Equation~\eqref{eq:state_transition}, and
updates the residual budget carried forward.  Dynamic programming exploits this structure by propagating value functions backward in time, while RL estimates those values on-line.  The temporal factorization reduces the dimensionality of each decision step from \(\mathcal{O}(2^{nh})\) to \(\mathcal{O}(2^{n})\). In the absence of a shared resource constraint or any other dependency which makes the component-level actions dependent, an additional decomposition is straightforward.  Because each asset $i$ evolves according to its own transition model in Equation~\eqref{eq:state_transition} and its reward
depends only on its own state and action, the global MDP factorizes into $n$ independent asset-level MDPs.  Each of these one-dimensional sub-problems can be solved by classical dynamic programming or by a solving separate RL problems, and the resulting per-asset policies can be executed in parallel. The picture changes once an annual budget constraint is introduced.
The requirement couples all asset actions at time $t$ as asset~$i$ selects an expensive intervention, fewer resources remain for the others.  This coupling breaks the asset-level independence, so the naive decomposition no longer yields a feasible global policy.

\subsubsection{Hierarchical task decomposition}
Hierarchical reinforcement learning inserts a higher-level controller that allocates resources or selects sub-tasks, leaving lower-level agents to plan detailed interventions. Many research have formulate the problem as a sequential decision making problem using MDP. Top-down MDP approaches, initially proposed by \citet{Golabi1982}, use linear programs that recommend the proportion of network assets to treat in each state and year, but they produce only randomized policies that must later be mapped to individual facilities.   Variants of this approach can be seen for bridge management \citet{golabi1997pontis} and its extensions by other researchers \citep{smilowitz2000optimal,kuhn2005model,madanat2006adaptive}.

By contrast, bottom-up approaches  \citet{Sathaye2012} optimize each facility separately, then resolve budget allocation each year to fit the system-wide constraint. Neither framework is fully satisfactory for large networks where future-year budget limits and interdependencies matter. Following the same approach,  \citet{zeng2024improved} replace the knapsack stage with a genetic algorithm to explore a larger combinatorial space.

Researchers have sought hybrid solutions, such as the Simultaneous Network Optimization (SNO) framework of \citet{Medury2014}, in which the current year’s decisions are facility-specific while future actions remain randomized, ensuring that network-level budget constraints are systematically enforced. The superiority of that SNO approach over purely bottom-up plans has been shown in scenarios where budgets are tight and facility condition distributions can deviate from expected values over time.

\subsubsection{Component–wise MDP decomposition}
Multi‑agent RL (MARL) frameworks assign one
agent per asset and approximate the global action–value function as a
combination of local utilities, enabling \emph{decentralised execution with
centralised training}.  Early work such as
Value‑Decomposition Networks (VDN) \citep{sunehag2017value} sums individual
$Q_i$’s, whereas QMIX \citep{rashid2020monotonic} uses a monotonic mixing network
to capture limited inter‑asset interaction and was applied by
Yao \citep{Yao2024} to lane‑level pavement maintenance under annual budgets.
QTRAN \citep{son2019qtran}, CMIX \citep{liu2021cmix} and
FACMAC \citep{peng2021facmac} relax the monotonicity assumption, while
mean‑field MARL \citep{mondal2024mean} replaces the exact joint policy with its
population average, yielding complexity that is linear in the number of
assets.  Although these factorisations greatly improve scalability, most
variants rely on Lagrangian penalties to impose resource ceilings; the recent
constraint‑aware CMIX shows promise, but Chen’s hardness result \citep{chen2024hardness} warns that a duality gap may persist in general.

\subsection{Constraint-handling techniques in constrained MDPs}

Multi-year budget planning falls naturally under the framework of
constrained Markov Decision Processes (CMDPs) \citep{altman2021constrained}, in which an agent maximises a primary return while satisfying one or more cost constraints.  Four broad families of constraint-handling methods appear in the RL and IAM literatures.

\subsubsection{Reward shaping with fixed penalties.}
The simplest approach augments the primary reward with a large negative penalty whenever a budget limit is breached.  \citet{Lei2022} embed the total life‑cycle budget in the reward of a DQN that schedules 200 bridges; \citet{Asghari2023} follow the same idea in a multi‑agent setting (DQN and A2C) by subtracting a fixed cost whenever the annual bridge budget is exceeded.  Although easy to implement, soft penalties offer no feasibility guarantee: the learned policy can still violate constraints if the penalty is not tuned carefully, and much of the exploration occurs in infeasible regions, slowing convergence.

\subsubsection{Lagrangian relaxation and primal–dual learning}
A principled way to enforce constraints is to dualise them, turning the CMDP into an unconstrained problem with modified reward $r_t-\sum_j\lambda_j c^j_t$ and updating the multipliers
$\lambda_j$ alongside the policy.
\citet{Caramanis2014} demonstrate that, under strong duality, ellipsoid‑type column‑generation can solve large CMDPs with complexity similar to value iteration.  In an IAM context, \citet{Faddoul2013} split an interdependent multi‑structure problem into tractable sub‑MDPs by relaxing the annual budget.  \citet{Andriotis2021} apply a primal–dual actor–critic inside a multi‑agent POMDP, updating $\lambda$ online to respect both annual and life‑cycle budgets.  The main caveat, revealed by the hardness analysis of \citet{chen2024hardness}, is that in cooperative MARL the feasible policy set is non‑convex; a non‑zero duality gap may therefore prevent the primal–dual iteration from reaching global optimality or even from converging.

\subsubsection{State augmentation and hard budget dynamics}
An alternative is to embed the remaining budget directly in the state so that any action leading to overspend is rendered infeasible.  Andriotis \citet{Andriotis2021} augment each agent’s observation with the residual annual and cumulative budgets, thereby guaranteeing feasibility by construction.  \citet{Yao2024} combine this idea with QMIX in pavement management, yet still allow occasional violations that are discouraged by a penalty term; hence budget feasibility is hard only if the penalty is set sufficiently high, which once more increases exploration inefficiency.

\subsubsection{Hierarchical or decomposition‑based schemes}
Finally, budget compliance can be enforced by a higher‑level controller that allocates resources, while a lower‑level policy decides individual interventions.  Hamida \citep{Hamida2023} and \citet{Tseremoglou2024} employ hierarchical DQN structures, but their top layer still relies on penalty shaping.  The multi‑agent approach of \citet{Yao2024} decomposes at the pavement‑segment level and aggregates costs through a mixing network, yet feasibility remains approximate.  In the next section we build on this line of work and propose a two‑level
\emph{hierarchical deep RL} (HDRL) architecture in which the upper actor outputs a continuous budget fraction and the lower actor selects interventions through a knapsack layer, guaranteeing exact satisfaction of both annual and life‑cycle constraints while keeping the action dimension linear in the number of assets.

\subsection{Concluding Remarks}
In addition to the scalability and constraints handling, substantial progress has been made in state representation, action space abstraction, sequential decision-making under uncertainty, and so on. Despite these advancements, critical research gaps and practical limitations persist.

Firstly, scalability remains a central challenge. Multi-agent reinforcement learning (MARL) frameworks, such as VDN, QMIX, CMIX, and FACMAC, have significantly improved scalability by decomposing complex decision-making tasks into decentralized execution supported by centralized training strategies \citep{sunehag2017value,rashid2020monotonic,liu2021cmix,peng2021facmac}. However, these methods face inherent limitations stemming from intricate inter-agent dependencies and persistent duality gaps associated with constraint management \citep{chen2024hardness}. Consequently, effectively managing large asset networks still necessitates advanced decomposition and coordination mechanisms.

Secondly, exact adherence to budget constraints remains a prominent unresolved issue. Simple methods relying on reward shaping or penalty terms, although straightforward to implement, fail to guarantee rigorous feasibility, often resulting in budget violations and inefficient exploration patterns \citep{Lei2022,Asghari2023}. More sophisticated approaches, including primal-dual methods and state augmentation techniques, offer improved constraint compliance but struggle with convergence stability, computational complexity, and compromise between accuracy and scalability \citep{Caramanis2014,Andriotis2021,Yao2024}.

Additionally, interpretability and transparency of RL-driven policies represent a relatively under-explored area. Most RL approaches, especially those using complex neural architectures or hierarchical decision-making frameworks, yield decision policies that are difficult to interpret intuitively, posing barriers to trust and adoption by infrastructure managers and decision-makers.

Moreover, stable training under strict budget constraints continues to be an open and challenging problem. Training RL algorithms within constrained, multi-year environments frequently results in unstable or slow convergence. Primal-dual iteration methods, for example, are susceptible to oscillations or suboptimal convergence due to non-convex feasible sets and duality gaps \citep{chen2024hardness}.

%These limitations underscore the urgent need for an integrated framework that effectively combines hierarchical decomposition, explicit constraint handling, and scalable reinforcement learning strategies. Such a unified framework should provide robust feasibility assurances, stable training convergence, enhanced interpretability, and effective scalability, significantly advancing multi-year IAM optimization capabilities.

Our proposed framework explicitly addresses these challenges by integrating three core components: a scalar Budget Planner, a priority-vector Maintenance Planner, and a linear-programming (LP) projection layer, all embedded within an off-policy Soft Actor–Critic (SAC) backbone. This innovative design advances the current state-of-the-art in several critical ways. Firstly, it introduces a one-dimensional budget allocation action, decoupling temporal budget decisions from detailed asset-level choices and thus avoiding exponential action-space growth. Secondly, it employs a knapsack-style LP projection to translate priority scores into explicit maintenance decisions, ensuring strict and exact compliance with both annual and lifecycle budget constraints. Thirdly, the hierarchical SAC structure achieves a linear action dimensionality of , significantly enhancing stable off-policy learning performance for larger-scale networks than previously possible with existing hierarchical or multi-agent reinforcement learning approaches.

\section{Key Concepts of Reinforcement Learning}\label{sec:rl-basics}

Reinforcement Learning (RL) provides a flexible framework for sequential decision-making under uncertainty, casting problems as Markov Decision Processes (MDPs). Instead of prescribing actions a priori, RL allows an agent to learn a policy by interacting with an environment, receiving states and rewards, and refining its decisions based on long-run outcomes. In infrastructure asset management, the environment can be viewed as a simulation of facilities deteriorating over time, where the agent chooses maintenance, rehabilitation, or replacement actions. By repeatedly testing different action sequences and observing both the immediate costs and cumulative benefits, the agent can discover strategies that optimize resource usage over multiple years.

An RL agent typically operates in discrete steps \(t = 0,1,2,\dots,T\). At step \(t\), the agent observes a state \(s_t \in \mathcal{S}\), chooses an action \(a_t \in \mathcal{A}\), and then observes a scalar reward \(r(s_t,a_t)\). The next state \(s_{t+1}\) is determined by transition dynamics \(p(s_{t+1}\mid s_t,a_t)\). Over repeated interactions, the agent refines a policy \(\pi\), mapping states to actions (or distributions over actions), with the goal of maximizing cumulative reward.

A widely used performance measure is the discounted return:
\begin{equation}\label{eq:return}
  G_t 
  \;=\; 
  \sum_{k=0}^{T - t - 1} 
    \gamma^k 
    \, r\bigl(s_{t+k}, a_{t+k}\bigr),
\end{equation}
where \(0 \le \gamma < 1\) is a discount factor. The action-value function \(Q_{\pi}(s,a)\) captures the expected return starting in state \(s\), taking action \(a\), and then following policy \(\pi\):
\begin{equation}\label{eq:qfunc}
  Q^{\pi}(s,a) 
  \;=\; 
  \mathbb{E}_{\pi}\!\Bigl[G_t 
  \,\Big|\,
  s_t = s,\; a_t = a\Bigr].
\end{equation}
Under Bellman’s principle of optimality, an optimal Q-function \(Q^*\) satisfies:
\begin{equation}\label{eq:qopt}
  Q^*(s,a) 
  \;=\; 
  r(s,a) \;+\;
  \gamma \sum_{s'} p\bigl(s'\mid s,a\bigr)\,
             \max_{a'}\,Q^*(s', a'),
\end{equation}
and yields an optimal policy:
\begin{equation}
      \pi^*(s)
  \;=\;
  \arg\max_{a\in\mathcal{A}}\,
  Q^*(s,a).
\end{equation}

In principle, solving for \(Q^*\) can be done by dynamic programming if \(\mathcal{S}\) and \(\mathcal{A}\) are finite and not too large. Real-world infrastructure problems, however, typically feature vast state-action spaces, especially once different condition states, budget constraints, and multi-year time horizons are included, necessitating approximate methods.

Q-learning \citep{sutton2018reinforcement} is a classic off-policy algorithm that incrementally approximates \(Q^*\). It maintains Q-values and updates them using a temporal-difference (TD) rule:
\begin{equation}\label{eq:qlearning}
  Q(s_t,a_t)
  \;\gets\;
  Q(s_t,a_t)
  \;+\;
  \alpha\,\Bigl(
    r(s_t,a_t) 
    \;+\;
    \gamma\,\max_{a'}\,Q\bigl(s_{t+1},a'\bigr)
    \;-\;
    Q(s_t,a_t)
  \Bigr),
\end{equation}
where \(\alpha\) is the learning rate. Over sufficient exploration, Q-learning converges to \(Q^*\) in problems with discrete, manageable state-action spaces and \(\gamma<1\).

However, Q-learning quickly becomes infeasible if \(\mathcal{S}\times\mathcal{A}\) is very large. Infrastructure management often has thousands of assets, each with multiple condition states and intervention options, leading to a combinatorial explosion in possible actions. Maintaining a Q-value table or straightforward function approximator for all \((s,a)\) pairs becomes impractical, prompting the development deep reinforcement learning (DRL) algorithms that use neural networks to generalize over large or continuous input spaces.

Deep Reinforcement Learning (DRL) integrates RL with deep neural networks to approximate value functions or policies, enabling the agent to handle high-dimensional states and actions. For instance, Deep Q-Learning (DQL) \citep{Mnih2013} replaces the Q-table with a neural network \(Q(s,a;\boldsymbol{\theta})\). The network takes a state \(s\) as input and outputs estimates of \(Q(s,a)\) for all valid actions \(a\). Training proceeds by fitting the network to a TD target:
\begin{equation}\label{eq:dqnloss}
  y_t
  \;=\;
  r(s_t,a_t) 
  \;+\;
  \gamma\,
  \max_{a'}\,
  Q\bigl(s_{t+1},\,a';\,\boldsymbol{\theta}\bigr),
\end{equation}
and minimizing the mean-squared error between \(y_t\) and \(Q(s_t,a_t;\boldsymbol{\theta})\). Techniques like Experience Replay (to break correlation in training samples) and Target Networks (to stabilize the TD target) are crucial to successful DQL training.

Despite these advances, DRL still faces challenges in infrastructure planning. When each year’s action is combinatorial (e.g., treat/do-nothing decisions for each of \(n\) assets) and subject to budget constraints, enumerating all feasible actions leads to exponentially large output layers, even if the environment’s state dimension is moderate. Specialized architectures, such as hierarchical RL or actor-critic with embedded optimization steps, are often needed to handle these budget-coupled decisions efficiently.

Whereas DQL and Q-learning are value-based, policy gradient approaches directly optimize a parameterized policy \(\pi_{\boldsymbol{\theta}}(a\mid s)\) via gradient ascent \citep{schulman2015high}. The expected return under \(\pi_{\boldsymbol{\theta}}\) can be written as:
\begin{equation}
    J\bigl(\pi_{\boldsymbol{\theta}}\bigr)
  =
  \mathbb{E}_{s\sim p,\,a\sim \pi_{\boldsymbol{\theta}}}\bigl[G_t\bigr],
\end{equation}
and the policy gradient theorem shows:
\begin{equation}
    \nabla_{\boldsymbol{\theta}}\,J\bigl(\pi_{\boldsymbol{\theta}}\bigr)
  =
  \mathbb{E}_{s,a}\Bigl[
    \nabla_{\boldsymbol{\theta}}\log \pi_{\boldsymbol{\theta}}(a\!\mid\!s)
    \,\bigl(G_t - b(s)\bigr)
  \Bigr],
\end{equation}
where \(b(s)\) is a baseline (often a state-value function) to reduce variance in the gradient estimate. Algorithms like REINFORCE or actor-critic variants (e.g., A2C, PPO) use Monte Carlo or bootstrapping techniques to estimate \(G_t\) and update \(\boldsymbol{\theta}\) accordingly. In contrast to value-based methods that derive a policy by greedily acting on learned Q-values, policy gradient methods learn the policy parameters directly, which can be more stable in large or continuous action spaces.

Soft Actor-Critic (SAC) \citep{haarnoja2018soft} is an actor-critic algorithm designed for continuous actions. It adds an entropy bonus to the reward, encouraging the agent to maintain a sufficiently stochastic policy, especially early in training. Formally, each transition’s reward is augmented by \(\alpha\,H(\pi_{\boldsymbol{\theta}}(\cdot\mid s))\), where \(\alpha\) is a temperature parameter. This entropy term prevents premature convergence to a deterministic strategy and fosters better exploration. 

In practice, SAC employs \textit{twin Q-networks}, two separate neural networks that estimate Q-values to reduce overestimation bias. A separate actor network \(\pi_{\boldsymbol{\theta}}\) is updated by maximizing the following:
\begin{equation}
  \mathbb{E}\Bigl[\min_{j=1,2}Q_{\phi_j}(s,\tilde{a}_{\boldsymbol{\theta}}(s,\epsilon))
  \;-\;
  \alpha\log\pi_{\boldsymbol{\theta}}(\tilde{a}_{\boldsymbol{\theta}}(s,\epsilon)\mid s)\Bigr],  
\end{equation}
where \(\tilde{a}_{\boldsymbol{\theta}}(s,\epsilon)\) is a reparameterized sample from the actor. Because SAC is off-policy, it can reuse data from a replay buffer, boosting sample efficiency. The entropy-regularized objective helps preserve exploratory diversity in high-dimensional scenarios, making SAC well suited to infrastructure problems where each decision has long-term consequences, and local myopia can be costly.

In many RL applications, each action is chosen from a relatively small discrete or continuous set. However, infrastructure asset management typically introduces budget constraints, linking what happens to each asset over the planning horizon. When \(n\) assets must each be either maintained or not, the naïve action space is on the order of \(\mathcal{O}(2^n)\) for each time step, which becomes unmanageable for even moderate \(n\). Additional constraints, such as maximum allowable annual/cumulative expenditures, compound the difficulty. 

While standard DRL techniques can theoretically incorporate constraints through reward shaping or penalty terms, they often fail to scale gracefully when action spaces are combinatorial. This has motivated hierarchical RL approaches, in which one policy allocates budgets, and another chooses which assets to treat within that budget. Techniques like Soft Actor-Critic are well suited here because they naturally handle continuous resource-allocation decisions and can integrate local subroutines (e.g.\ knapsack solvers) for the discrete maintenance decisions. Section~\ref{sec:HDRL} presents a Hierarchical Deep Reinforcement Learning (HDRL) method explicitly tailored for these budget-coupled, combinatorial actions.

\medskip

In summary, RL rests on the principle of repeated state-action interactions, guided by a policy that seeks to maximize long-run rewards. Q-learning, policy gradients, and hybrid actor-critic methods each offer different ways of approximating or iterating toward optimal policies. However, budget constraints and exponentially large action sets pose additional challenges in the infrastructure domain. The next section addresses these complexities through a hierarchical decomposition that sidesteps enumerating every feasible subset, allowing RL agents to operate effectively even in large, resource-limited systems.

\section{Proposed Method: Hierarchical Deep Reinforcement Learning (HDRL)}\label{sec:HDRL}

This section presents a Hierarchical Deep Reinforcement Learning (HDRL) framework designed to handle multi-year maintenance decisions under strict budget constraints. The core concept is to decompose each year’s decision into two levels: (1) how much of the remaining budget to allocate for the current year, and (2) which maintenance actions to execute given that allocated budget. This two-tier process enables more tractable learning than a single-agent RL approach would offer, particularly when the environment has combinatorial state and action spaces.

\subsection{Decomposing the Annual Decisions into Two Levels}

Consider a planning horizon of \(h\) years and \(n\) assets. At each decision epoch~\(t\), two sequential actions are taken. First, Actor 1 (the Budget Planner) determines a scalar \(a_{t}^{(1)} \in [0,1]\). This fraction represents what portion of the remaining multi-year budget should be allocated to year~\(t\). This fraction is mapped to an admissible annual budget \(b_{t}\) while enforcing constraints on yearly expenditures:
\begin{equation}\label{eq:annualBudgetMapping}
b_{t}
=\;
\max\bigl(
b_{t}^{l} \,+\, (a_{t}^{(1)}+1)\,\bigl(b_{t}^{u} - b_{t}^{l}\bigr)/2,
\;
b^{\text{total}}
-\!\sum_{k=1}^{t-1} b_{k}
-\!\sum_{k=t+1}^{h} b_{k}^{l}
\bigr).
\end{equation}

This ensures at least a minimum budget is allocated to future years. Actor~1’s goal is to learn an effective distribution of spending over time, so as to maximize overall long-run performance.

At the second level, Actor 2 (the Maintenance Planner)  receives the chosen annual budget action ( \(a_{t}^{(1)}\)) and produces a vector of priority coefficients~\(\mathbf{a}_{t}^{(2)}\in\mathbb{R}^{n}\) for the \(n\) assets. Given these coefficients, a local linear program (LP) determines which assets to maintain in year~\(t\), ensuring total cost stays within \(b_{t}\). These coefficients are designed not merely to maximize next-year condition, but rather to reflect the multi-year impact of interventions.
Actor~2 outputs a vector of asset-level coefficients \(\mathbf{a}_{t}^{(2)}=\bigl(a_{1,t}^{(2)},\dots,a_{n,t}^{(2)}\bigr)\). These coefficients translate into a local linear program (LP) for choosing which assets to maintain that year:
\begin{align}
&\max_{\{x_{i,t}\}} \;\;
\sum_{i=1}^{n} a_{i,t}^{(2)}
\,\Bigl[f_{i}^{\mathrm{act}}(s_{i,t}) - f_{i}^{\mathrm{det}}(s_{i,t})\Bigr]
\,x_{i,t}, \label{eq:lp-maintenance}
\\[6pt]
&\text{subject to:}\quad
\sum_{i=1}^{n} c_{i,t}\,w_{i}\,x_{i,t}\;\leq\;b_{t},\quad
\sum_{i=1}^{n} c_{i,t}\,w_{i}\,x_{i,t}\;\ge\;b_{t}^{l},
\nonumber\\
&\qquad\qquad\quad x_{i,t}\in\{0,1\},\ \forall i.
\nonumber
\end{align}
Here,
\(\bigl[f_{i}^{\mathrm{act}}(s_{i,t}) - f_{i}^{\mathrm{det}}(s_{i,t})\bigr]\)
represents the immediate gain (e.g., next-year improvement) from intervening on asset~\(i\). 
When all coefficients \(a_{i,t}^{(2)}\) are equal 1, this LP simply maximizes next year’s condition gain subject to the annual budget. 
By adjusting \(a_{i,t}^{(2)}\), however, the Maintenance Planner weights each asset’s relative importance in a multi-year sense, effectively learning how to transform the one-year objective into one that maximizes the Level of Service of the remaining horizon.  
Hence, each year’s decision \(\mathbf{a}_{t}\) is split into \(\bigl[a_{t}^{(1)},\ \mathbf{a}_{t}^{(2)}\bigr]\).
By learning the budget fraction \(\bigl(a_{t}^{(1)}\bigr)\) and the asset-specific priorities \(\bigl(\mathbf{a}_{t}^{(2)}\bigr)\) separately, the HDRL agent avoids enumerating an exponential number of full-system actions in one step. By factorizing the problem into a scalar budget action plus an \(n\)-dimensional maintenance action, and further enforcing feasibility via a knapsack subproblem, HDRL substantially reduces the complexity compared to a monolithic RL agent (e.g. DQL) with a combinatorial action space of size \(2^n\) per step.

\begin{figure}[H]
  \centering
  \includegraphics[width=\linewidth]{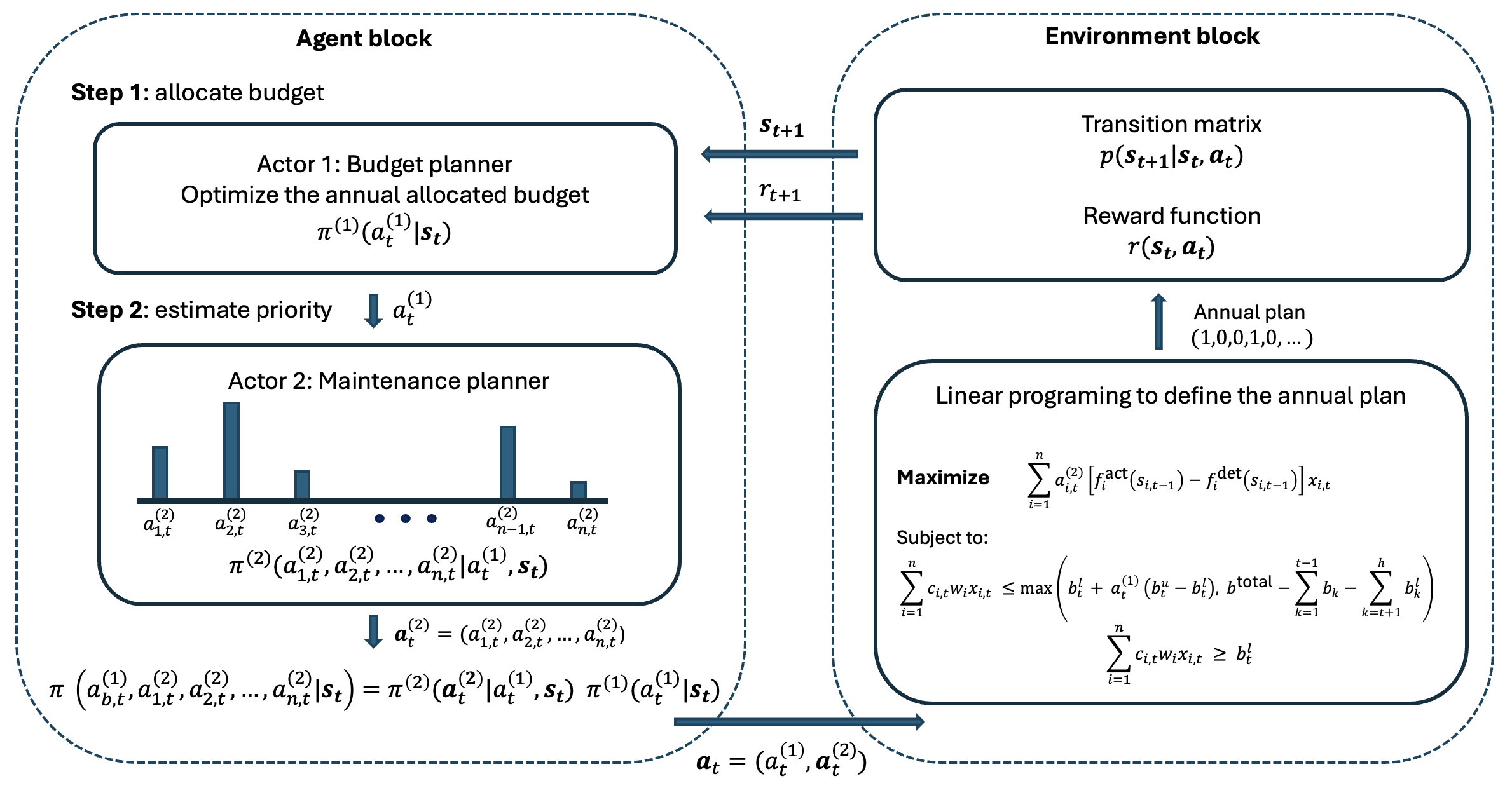}
  \caption{HDRL architecture with two-level decision decomposition. Actor~1 selects the annual budget fraction; Actor~2 computes priority scores for assets. A linear program determines the annual plan. The environment updates the state using a Markov transition model and computes the reward based on the selected actions.}
  \label{fig:hdrl}
\end{figure}

Figure~\ref{fig:hdrl} summarizes the full decision-making flow of the HDRL framework. The agent block consists of two steps: the Budget Planner first allocates the annual budget, and the Maintenance Planner then assigns maintenance priorities to assets. These priorities are translated into binary decisions through the LP block. The resulting annual plan is submitted to the environment block, which updates the system state using the transition function \(p(\mathbf{s}_{t+1} \mid \mathbf{s}_t, \mathbf{a}_t)\) and computes the corresponding reward using \(r(\mathbf{s}_t, \mathbf{a}_t)\). The next state and reward are returned to the agent, completing the loop.

By structuring the decision process in this way, the HDRL framework avoids the need to enumerate all possible combinations of asset-level interventions. Instead, it reduces the decision complexity from exponential in the number of assets to linear in the number of priority coefficients. This hierarchical factorization enables the model to scale to large infrastructure systems and ensures that budget constraints are respected at every step.

\subsection{State Vector, Reward and Objective}
The global state at time (or year) \(t\) is augmented with four key elements:
\begin{equation}
\mathbf{s}_{t} 
=
\Bigl[
\,s_{1,t},\,s_{2,t},\dots,s_{n,t},\;
\mathrm{LoS}_{t},\;
\tfrac{t}{h},\;
b_{t}^{r}
\Bigr].
\end{equation}
Here, each \(s_{i,t}\) represents the condition of asset \(i\) (possibly a probability distribution or a discrete index). The term \(\frac{t}{h}\) encodes normalized time, allowing the agent to differentiate early- vs.\ late-horizon decisions. Finally, 
\begin{equation}
b_{t}^{r}
=\;
\frac{b^{\text{total}} - \sum_{k=1}^{t-1} b_{k}}{\,b^{\text{total}}\,}
\end{equation}
tracks the ratio of remaining budget after \(t-1\) years, with \(b^{\text{total}}\) denoting the total permissible expenditure across the entire horizon, and \(b_{k}\) the budget used in year \(k\). This augmented state thus merges asset-level, global performance, temporal, and budget information in one vector, ensuring the RL agent can learn intertemporal and system-wide strategies.

Upon applying \(\bigl(a_{t}^{(1)},\mathbf{a}_{t}^{(2)}\bigr)\) and the resulting annual plan, the states being updated, and the environment yields a scalar reward \(r_{t}\). The reward function should be designed in a way that the cumulative reward over the horizon would be equivalent to the objective function of the original optimization problem. 
%
%The reward is the next year level of service \(\mathrm{LoS}_{t+1}\), namely, $r_{t} = \mathrm{LoS}_{t+1}$. 
%
The planning objective in Equation~\eqref{eq:objective} maximises the horizon-average level of service, namely,
\(\ \left(\sum_{t=1}^{h}\mathrm{LoS}_{t} \right)/h\).
Since the episode length is \(h\) years, we define the
one-step reward as
\(r_t\!=\!\mathrm{LoS}_{t+1}\).
Therefore, the undiscounted episodic return is 
\begin{equation}\label{eq:eqv_return}
  G_0
  =\sum_{t=0}^{h-1} r_t
  =\sum_{t=0}^{h-1}\mathrm{LoS}_{t+1}
  =\sum_{t=1}^{h}\mathrm{LoS}_{t},
\end{equation}
which is proportional to the horizon-average objective.

% The annual budget allocation mechanism represented by Equation (\ref{eq:annualBudgetMapping}) is designed to avoid violation of annual and total budget constraints. However additional penalty terms can be added to the reward if needed to enforce a specific constraint (e.g. performance limits).

\subsection{Neural Network Architecture and Learning Process}
\label{sec:HDRL:NN-full}

\begin{figure}[H]
  \centering
  \begin{subfigure}{\linewidth}
    \centering
    \includegraphics[width=0.75\linewidth]{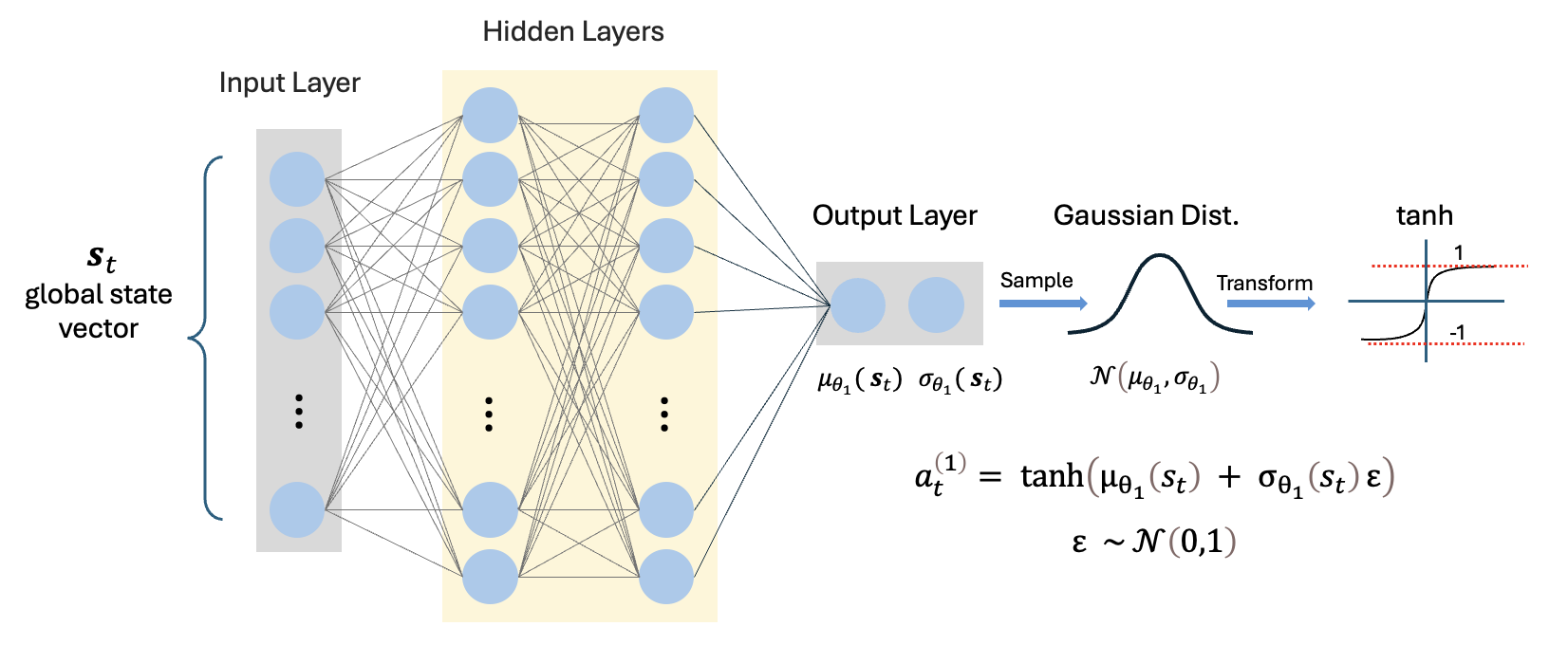}
    \caption{Budget Planner (Actor 1)}
  \end{subfigure}

  \begin{subfigure}{\linewidth}
    \centering
    \includegraphics[width=0.75\linewidth]{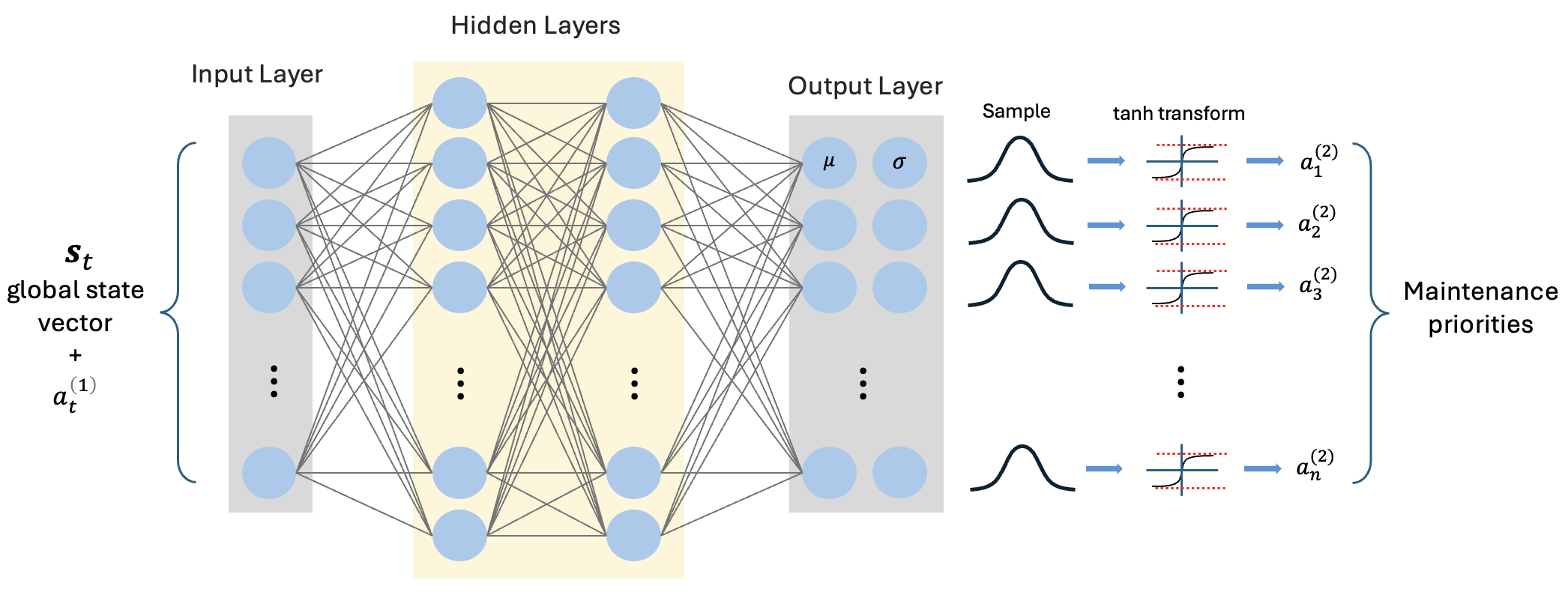}
    \caption{Maintenance Planner (Actor 2)}
  \end{subfigure}

  \begin{subfigure}{\linewidth}
    \centering
    \includegraphics[width=0.50\linewidth]{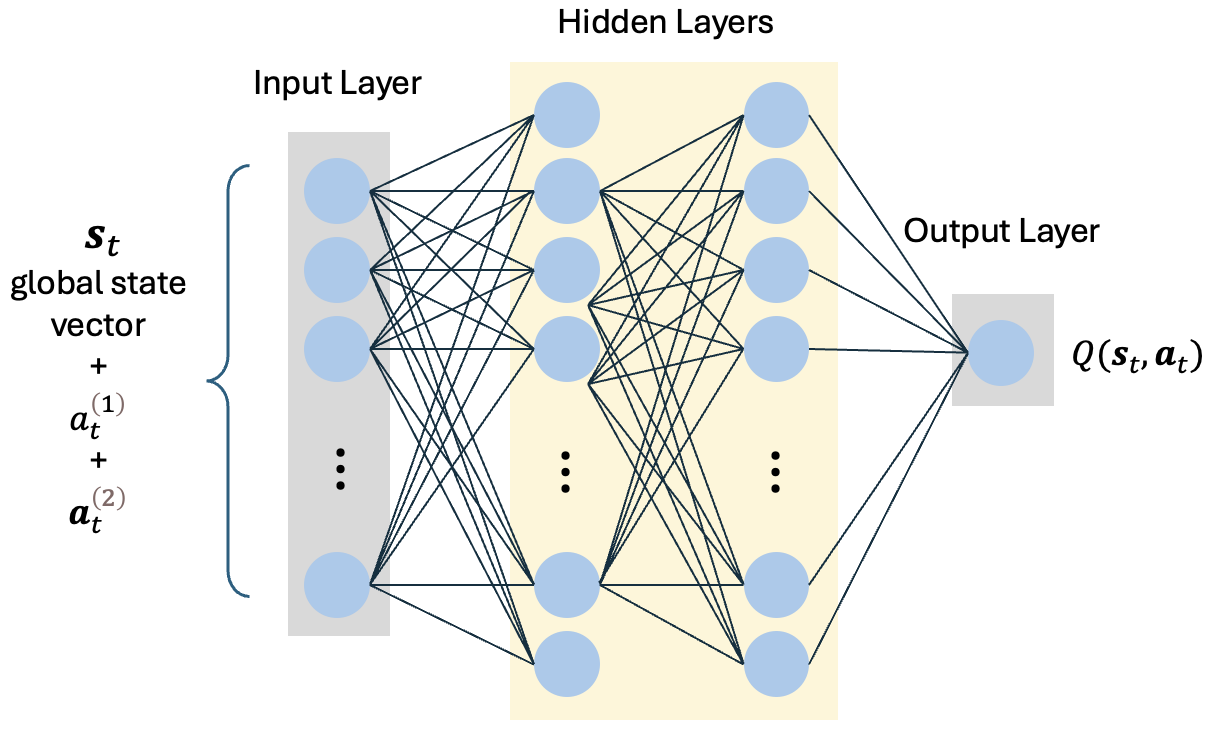}
    \caption{Shared topology of the twin soft-Q critics}
  \end{subfigure}

  \caption{Neural-network architectures used in the hierarchical SAC framework. (a) and (b) depict the two stochastic
           actors; (c) shows the critic network employed by both
           \(Q_{\phi_{1}}\) and \(Q_{\phi_{2}}\).}
  \label{fig:net_architectures}
\end{figure}

The proposed HDRL framework is composed of two actor networks and two critic networks, structured within the Soft Actor-Critic (SAC) paradigm (Figure \ref{fig:net_architectures}). These networks collaboratively learn policies for annual budget allocation and asset-level maintenance prioritization, while maintaining compliance with budget constraints. The SAC framework is especially well-suited for this hierarchical structure because it supports stable, sample-efficient off-policy learning and facilitates continuous exploration through entropy maximization.

As shown in Figure~\ref{fig:net_architectures}(a), Actor 1 or the \emph{Budget Planner} receives the global state vector \(\mathbf{s}_t\) as input and outputs a scalar action \(a_t^{(1)}\), which determines the proportion of the remaining budget to allocate in year \(t\). The input state vector includes asset condition profiles, the current year normalized by the planning horizon, system-level performance metrics, and the remaining budget ratio. The actor network includes two fully connected hidden layers with ReLU activations, followed by two output heads that represent the mean \(\mu_{\theta_1}(\mathbf{s}_t)\) and the log standard deviation \(\log \sigma_{\theta_1}(\mathbf{s}_t)\) of a Gaussian distribution. A noise vector \(\varepsilon \sim \mathcal{N}(0,1)\) is used in the reparameterization step to produce the action:
\begin{equation}
    a_t^{(1)} = \tanh\left( \mu_{\theta_1}(\mathbf{s}_t) + \sigma_{\theta_1}(\mathbf{s}_t) \cdot \varepsilon \right).
\end{equation}
The use of the \(\tanh\) function ensures that the output remains bounded in \([-1,1]\), and this value is subsequently mapped to a feasible budget allocation using Equation~\eqref{eq:annualBudgetMapping}.

The \emph{Maintenance Planner} or Actor 2 receives both \(\mathbf{s}_t\) and the sampled budget action \(a_t^{(1)}\), and outputs an \(n\)-dimensional vector \(\mathbf{a}_t^{(2)}\), where each element represents the priority score for a specific asset. As depicted in Figure~\ref{fig:net_architectures}(b), this actor network mirrors the Budget Planner in architecture, including two hidden layers and Gaussian policy heads. Each output dimension is independently sampled, transformed through a \(\tanh\) function, and passed to a binary knapsack linear program \eqref{eq:lp-maintenance} that selects a subset of assets for intervention. This local optimization ensures that all budget constraints are satisfied.

Sharing the same network topology as shown in Figure~\ref{fig:net_architectures}(c), the two \emph{critic networks}, \(Q_{\phi_1}\) and \(Q_{\phi_2}\), estimate the soft Q-values of state-action pairs. They receive the concatenated input \((\mathbf{s}_t, a_t^{(1)}, \mathbf{a}_t^{(2)})\), which is passed through two hidden layers to produce scalar Q-value estimates. The loss for each critic is computed using the soft Bellman error:
\begin{equation}
    \mathcal{L}_Q = \mathbb{E} \left[ \left( Q_{\phi_j}(\mathbf{s}_t, a_t^{(1)}, \mathbf{a}_t^{(2)}) - y_t \right)^2 \right],
\end{equation}
where the target value \(y_t\) is defined as:
\begin{equation}
    y_t = r_t + \gamma \left[ \min_j Q_{\phi_j}(\mathbf{s}_{t+1}, a_{t+1}^{(1)}, \mathbf{a}_{t+1}^{(2)}) - \alpha \mathcal{H}_{t+1} \right],
\end{equation}
and the entropy term \(\mathcal{H}_{t+1}\) is expressed:
\begin{equation}
    \mathcal{H}_{t+1} = \log \pi_{\theta_1}(a_{t+1}^{(1)} \mid \mathbf{s}_{t+1}) + \log \pi_{\theta_2}(\mathbf{a}_{t+1}^{(2)} \mid \mathbf{s}_{t+1}, a_{t+1}^{(1)}).
\end{equation}

Regarding the training of the networks, the \emph{actor networks} are optimized so as to maximize soft expected return. The loss function for Actor 1 (the Budget Planner) is given by:
\begin{equation}
\mathcal{L}_{\pi^{(1)}} = \mathbb{E} \left[
\alpha \log \pi_{\theta_1}(a_t^{(1)} \mid \mathbf{s}_t) - \min_j Q_{\phi_j}(\mathbf{s}_t, a_t^{(1)}, \mathbf{a}_t^{(2)})
\right],
\end{equation}
where \(\mathbf{a}_t^{(2)}\) is sampled from \(\pi_{\theta_2}(\cdot \mid \mathbf{s}_t, a_t^{(1)})\). For Actor 2 (Maintenance Planner), the loss function is similar:
\begin{equation}
\mathcal{L}_{\pi^{(2)}} = \mathbb{E} \left[
\alpha \log \pi_{\theta_2}(\mathbf{a}_t^{(2)} \mid \mathbf{s}_t, a_t^{(1)}) - \min_j Q_{\phi_j}(\mathbf{s}_t, a_t^{(1)}, \mathbf{a}_t^{(2)})
\right].
\end{equation}

The \emph{entropy temperature parameter} \(\alpha\) is a learnable scalar that regulates the trade-off between exploration and exploitation. A high value of \(\alpha\) encourages more stochastic behavior, whereas a low value leads to more deterministic policies. Following the entropy tuning method introduced by \citet{haarnoja2018soft}, \(\alpha\) is updated by minimizing the loss:
\begin{equation}
    \mathcal{L}_\alpha = \mathbb{E} \left[
-\alpha \left( \log \pi(a_t \mid \mathbf{s}_t) + \bar{\mathcal{H}} \right)
\right],
\end{equation}
where \(\bar{\mathcal{H}}\) is the target entropy. The gradient is given by:
\begin{equation}
    \nabla_\alpha \mathcal{L}_\alpha = -\left( \log \pi_{\theta_1}(a_t^{(1)} \mid \mathbf{s}_t) + \log \pi_{\theta_2}(\mathbf{a}_t^{(2)} \mid \mathbf{s}_t, a_t^{(1)}) + \bar{\mathcal{H}} \right),
\end{equation}
and thus the parameter is udpated as $  \alpha \leftarrow \alpha - \eta_\alpha \nabla_\alpha \mathcal{L}_\alpha$. To stabilize learning, each critic has an associated \emph{target network} \(\bar{Q}_{\phi_j}\), which is updated using Polyak averaging: $\bar{\phi}_j \leftarrow \tau \phi_j + (1 - \tau) \bar{\phi}_j$, where \(\tau\) is a small constant (e.g., 0.005) that controls the update rate.

\begin{algorithm}[H]
\caption{HDRL training loop with explicit loss definitions for actors, temperature, and target networks}
\label{alg:hdrl_train}
\small
\begin{algorithmic}[1]
\Require learning rates $\eta_{\phi},\eta_{\theta_1},\eta_{\theta_2},\eta_{\alpha}$,
         discount factor $\gamma$, Polyak factor $\tau$,  
         replay‑buffer capacity $C$, mini‑batch size $B$
\State Initialise actors $\pi_{\theta_1},\pi_{\theta_2}$ and critics $Q_{\phi_1},Q_{\phi_2}$ 
\State Set target critics $\bar Q_{\bar\phi_1}\!\leftarrow\! Q_{\phi_1}$ and
       $\bar Q_{\bar\phi_2}\!\leftarrow\! Q_{\phi_2}$
\State Create an empty replay buffer $\mathcal D$ with capacity $C$
\State Initialise entropy temperature $\alpha\leftarrow 0.1$
\For{each episode $e=1,\dots,N_{\text{ep}}$}
  \State Reset the environment and observe $\mathbf s_0$
  \For{year $t=0,\dots,h-1$}
     \State Sample budget fraction $a^{(1)}_{t}\!\sim\!\pi_{\theta_1}(\,\cdot\,|\,\mathbf s_t)$
     \State Sample priority vector $\mathbf a^{(2)}_{t}\!\sim\!
            \pi_{\theta_2}(\,\cdot\,|\,\mathbf s_t,a^{(1)}_{t})$
     \State Solve \eqref{eq:lp-maintenance} to obtain binary decision $\mathbf x_t$
     \State Execute $\mathbf x_t$, receive $r_t$ and next state $\mathbf s_{t+1}$
     \State Store $(\mathbf s_t,a^{(1)}_{t},\mathbf a^{(2)}_{t},r_t,\mathbf s_{t+1})$ in $\mathcal D$
     \If{$\lvert\mathcal D\rvert\ge B$}
       \State Sample mini‑batch $\mathcal B\subset\mathcal D$ of $B$ transitions
       \Statex\hspace*{1.5em}\textbf{Critic update}
       \State Compute target
       \[
       \begin{aligned}
       y_t
       &\,=\,r_t+\gamma\Bigl[
       \min_{k}\bar Q_{\bar\phi_k}
            (\mathbf s_{t+1},a^{(1)}_{t+1},\mathbf a^{(2)}_{t+1}) \\[-2pt]
       &\hphantom{=\;}
       -\alpha\bigl(
         \log\pi_{\theta_1}(a^{(1)}_{t+1}|\mathbf s_{t+1})
         +\log\pi_{\theta_2}(\mathbf a^{(2)}_{t+1}
                            |\mathbf s_{t+1},a^{(1)}_{t+1})
       \bigr)\Bigr]
       \end{aligned}
       \]

       \State Calculate Critic loss
              $\displaystyle
                 \mathcal L_{Q}=
                 \frac1B\!\sum_{(\mathbf s_t,\dots)\in\mathcal B}
                 \bigl(Q_{\phi_j}(\mathbf s_t,a^{(1)}_{t},\mathbf a^{(2)}_{t})-y_t\bigr)^2$
       \State Update \ $\phi_j$ using Adam  with $\eta_{\phi}$
       \Statex\hspace*{1.5em}\textbf{Actor\,1 (Budget) loss}
       \State $\displaystyle
              \mathcal L_{\pi^{(1)}}=
              \frac1B\!\sum_{(\mathbf s_t,\dots)\in\mathcal B}
              \bigl[
                \alpha\log\pi_{\theta_1}(a^{(1)}_{t}|\mathbf s_t)
                -\min_{k}Q_{\phi_k}(\mathbf s_t,a^{(1)}_{t},\mathbf a^{(2)}_{t})
              \bigr]$
       \State Update $\theta_1\leftarrow\theta_1-\eta_{\theta_1}\nabla_{\theta_1}\mathcal L_{\pi^{(1)}}$
       \Statex\hspace*{1.5em}\textbf{Actor\,2 (Maintenance) loss}
       \State $\displaystyle
              \mathcal L_{\pi^{(2)}}=
              \frac1B\!\sum_{(\mathbf s_t,\dots)\in\mathcal B}
              \bigl[
                \alpha\log\pi_{\theta_2}(\mathbf a^{(2)}_{t}|\mathbf s_t,a^{(1)}_{t})
                -\min_{k}Q_{\phi_k}(\mathbf s_t,a^{(1)}_{t},\mathbf a^{(2)}_{t})
              \bigr]$
       \State Update $\theta_2\leftarrow\theta_2-\eta_{\theta_2}\nabla_{\theta_2}\mathcal L_{\pi^{(2)}}$
       \Statex\hspace*{1.5em}\textbf{Temperature update}
       \State $\displaystyle
              \nabla_{\alpha}\mathcal L_{\alpha}=
              -\!\left(\!
                 \log\pi_{\theta_1}(a^{(1)}_{t}|\mathbf s_t)
                 +\log\pi_{\theta_2}(\mathbf a^{(2)}_{t}|\mathbf s_t,a^{(1)}_{t})
                 +\bar{\mathcal H}\!\right)$
       \State $\alpha\leftarrow\alpha-\eta_{\alpha}\,\nabla_{\alpha}\mathcal L_{\alpha}$
       \Statex\hspace*{1.5em}\textbf{Target‑network update}
       \State $\bar\phi_k\leftarrow
              \tau\,\phi_k+(1-\tau)\,\bar\phi_k,\quad k=1,2$
     \EndIf
  \EndFor
\EndFor
\end{algorithmic}
\end{algorithm}

The full training procedure is outlined in Algorithm~\ref{alg:hdrl_train}. Each episode simulates a planning horizon of \(h\) years. At every step \(t\), the actors sample actions, the environment returns the reward and next state, and the transition is added to a replay buffer. At regular intervals, a mini-batch of transitions is sampled to update all networks. The critic networks are trained using \(\mathcal{L}_Q\), the actors are updated via \(\mathcal{L}_{\pi^{(1)}}\) and \(\mathcal{L}_{\pi^{(2)}}\), and the temperature parameter is adjusted using \(\mathcal{L}_\alpha\). Finally, the target Q-networks are updated via Polyak averaging. This training loop is repeated over multiple episodes until convergence. The combined effect of entropy-regularized optimization, LP projection, and hierarchical decision decomposition ensures the agent learns robust and budget-compliant policies even in large-scale infrastructure networks with high-dimensional decision spaces.
       
\section{Case Study and Analysis}

This section compares the proposed Hierarchical Deep Reinforcement Learning (HDRL) method with a Deep Q-Learning (DQL) baseline on three progressively larger problems.  The experiments use a five-year planning horizon for a sewer network in an Ontario municipality \citep{Fard2024}.  Two budget constraints are enforced simultaneously.  First, a cumulative cap of \$500\,000 applies to the entire horizon.  Second, each year must respect both an upper and a lower spending limit derived from the average budget of \$100\,000: annual cost may vary by at most \(\pm5\%\), so the feasible range is between \$95\,000 and \$105\,000 in every year. 

Sewer segment condition is modelled on a discrete five-point scale, where 1 denotes pristine and 5 denotes failure.  Because lower numbers are preferable, the performance mapping $\kappa(s)=({5-s})/4$ converts a condition state to a 0–1 score that rises with better condition.  The level of service (LoS) for an individual sewershed equals the length-weighted average of its segment scores, and the network-wide LoS is the length-weighted average of sewershed scores.  Segment-level transition matrices are generated with the regression model of \citep{LinYuan19}, which relates deterioration probabilities to length, diameter, and slope.  Each sewershed’s transition matrix \(\boldsymbol{P}_i\) is then the length-weighted average of its segments’ matrices.  Descriptive statistics for the ten-sewershed base case appear in Table \ref{table:sewershed_data}; the instances with 15 and 20 sewersheds are formed by appending additional areas drawn from the same municipality. Flushing an asset resets its condition to prime, costing \$3 per unit length. 

%% The sewershed state \(\boldsymbol{s}_{i,t}\) evolves according to
%\[
%\boldsymbol{s}_{i,t}
%=
%(1 - x_{i,t})\,\bigl(\boldsymbol{s}_{i,t-1}\,\boldsymbol{P}_i\bigr)
%+
%x_{i,t}\,\bigl(\boldsymbol{s}_{i,t-1}\,\boldsymbol{P}_i\,\boldsymbol{M}_i\bigr),
%\]
%where \(x_{i,t}\in\{0,1\}\) indicates whether sewershed \(i\) is flushed at year \(t\).

\begin{table}[ht]
\centering
\caption{Details of the Ten Sewersheds}
\label{table:sewershed_data}
\small
\renewcommand{\arraystretch}{1.25}
\begin{tabular}{c c c c}
\hline\hline
Sewershed & \# of Segments & Total Length & Initial Avg. Condition \\
\hline
1 & 520 & 34{,}643.29 & 2.367 \\
2 & 463 & 30{,}783.54 & 1.346 \\
3 & 195 & 12{,}934.21 & 1.434 \\
4 & 193 & 11{,}667.71 & 1.312 \\
5 & 156 & 11{,}423.13 & 1.307 \\
6 & 137 & 11{,}393.38 & 1.493 \\
7 & 97  & 8{,}479.24  & 1.344 \\
8 & 96  & 7{,}359.18  & 1.000 \\
9 & 78  & 6{,}411.50  & 1.327 \\
10 & 89 & 5{,}939.84  & 1.388 \\
\hline\hline
\end{tabular}
\normalsize
\end{table}

\subsection{Baseline Validation: 10-Sewershed Case Study}

\subsubsection{Optimal Reference Solution for the 10-Sewershed Network}
A constraint programming (CP) model using OR-Tools \citep{ortools} and CP-SAT Solver \citep{cpsatlp} enumerates every flush/no-flush schedule over a five-year horizon for 10 sewersheds. Of the theoretical \(2^{50}\) possible plans, approximately 2{,}249{,}947 are budget-feasible. Table~\ref{table:optimal_solution} reports the optimal solution, with a total cost near \$498{,}925 and an objective value of 1.4687 \citep{Fard2024}. This plan is treated as a reference baseline (depicted as a black star in plots).

\begin{table}[ht]
\centering
\caption{Five-Year Optimal Flushing Plan for 10 Sewersheds}
\label{table:optimal_solution}
\small
\renewcommand{\arraystretch}{1.25}
\begin{tabular}{c|c|c|c|c|c}
\hline\hline
Sewershed & Year 1 & Year 2 & Year 3 & Year 4 & Year 5 \\
\hline
1 & 1 & 0 & 0 & 1 & 0 \\
2 & 0 & 0 & 0 & 0 & 0 \\
3 & 0 & 0 & 1 & 0 & 1 \\
4 & 0 & 0 & 1 & 0 & 0 \\
5 & 0 & 0 & 0 & 0 & 1 \\
6 & 0 & 1 & 0 & 0 & 0 \\
7 & 0 & 1 & 0 & 0 & 1 \\
8 & 0 & 0 & 1 & 0 & 0 \\
9 & 0 & 1 & 0 & 0 & 0 \\
10 & 0 & 1 & 0 & 0 & 0 \\
\hline
Total Length (km) & 34.64 & 32.22 & 31.96 & 34.64 & 32.84 \\
Annual Cost & 103{,}929.9 & 96{,}671.9 & 95{,}883.3 & 103{,}929.9 & 98{,}509.7 \\
\hline\hline
\end{tabular}
\normalsize
\end{table}

\subsubsection{DQL with Combinatorial Actions}
The naive action space at each decision step includes \(2^{10} = 1024\) flush/no-flush combinations. Imposing a budget constraint narrows this to only 20 feasible actions, so the last layer of the Q-network has size 20. Restricting outputs to those 20 actions avoids extra penalty terms in the reward and mitigates variance issues in training. A learning rate of \(8 \times 10^{-5}\) is used for the Q-network, which has two 256-node hidden layers. A prioritized replay buffer is applied (batch size 256). Over 5{,}000 episodes, training takes approximately 118 seconds on average (across 100 runs).

\subsubsection{HDRL Setting}
\begin{figure}[H]
    \centering
    \begin{subfigure}[b]{0.48\textwidth}
        \includegraphics[width=\linewidth]{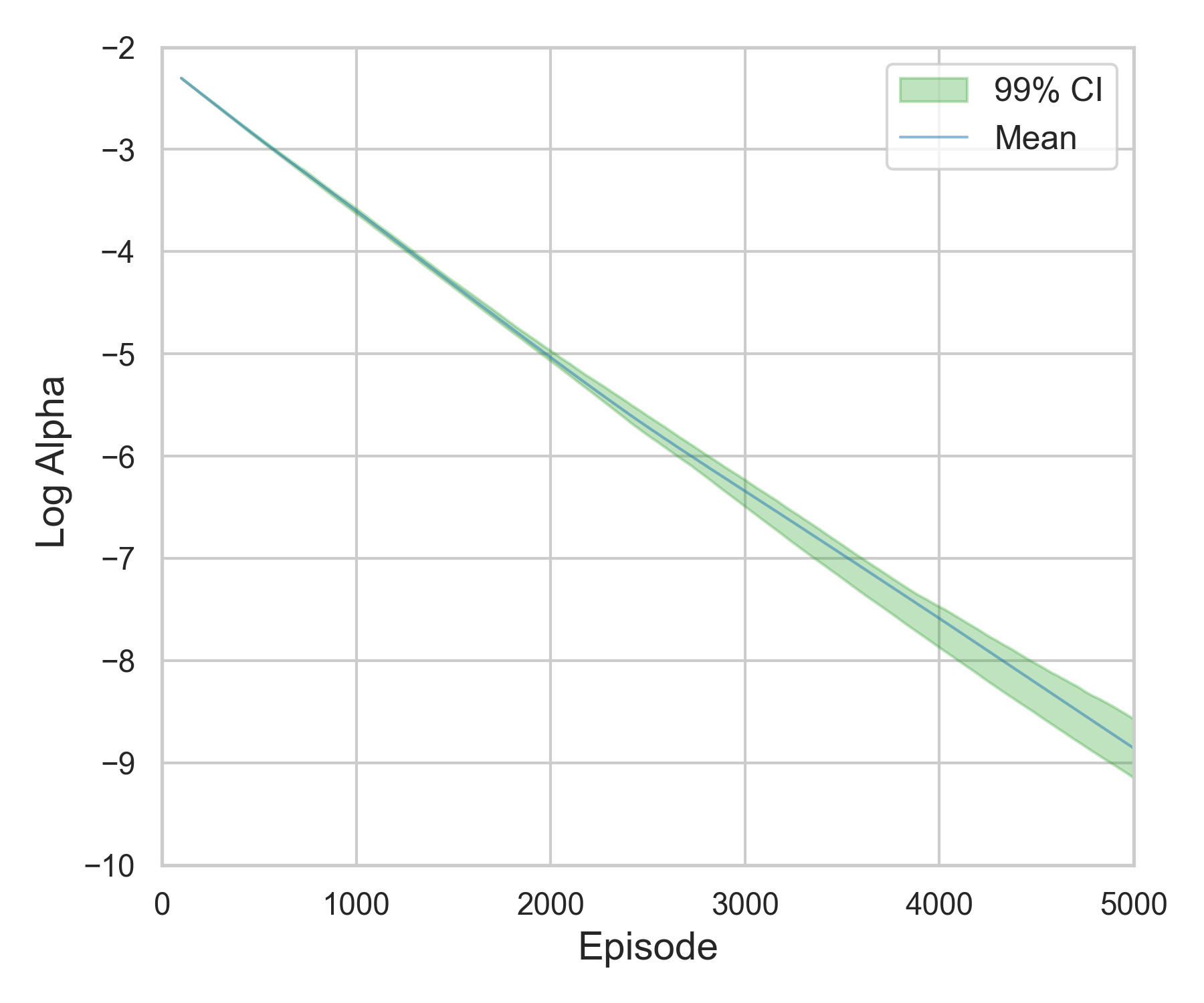}
        \caption{Entropy coefficient (\(\alpha\))}
        \label{fig:alpha}
    \end{subfigure}
    \hfill
    \begin{subfigure}[b]{0.48\textwidth}
        \includegraphics[width=\linewidth]{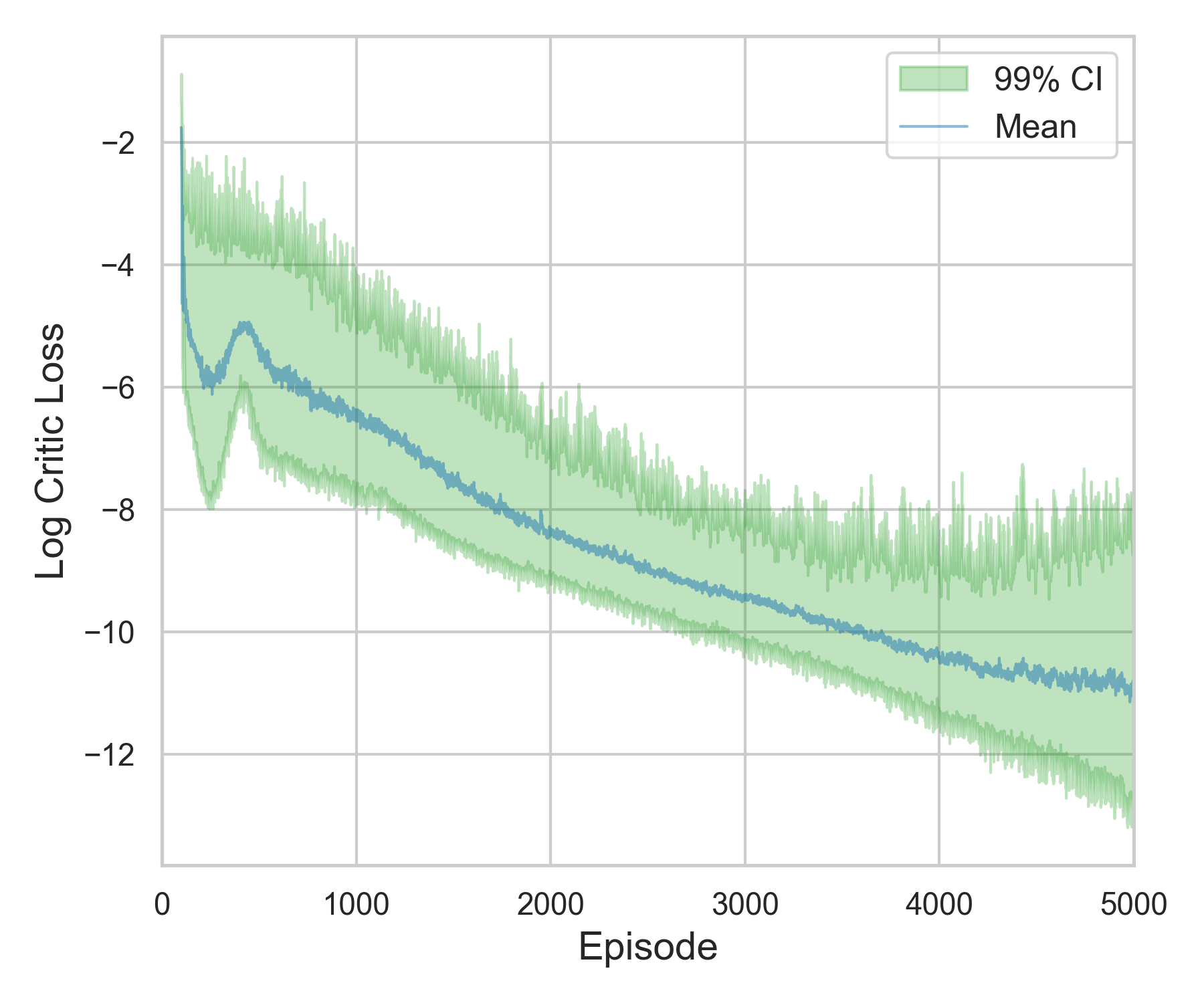}
        \caption{Max Critic Loss}
        \label{fig:critic_max_loss}
    \end{subfigure}
    
    \vspace{1em}
    
    \begin{subfigure}[b]{0.48\textwidth}
        \includegraphics[width=\linewidth]{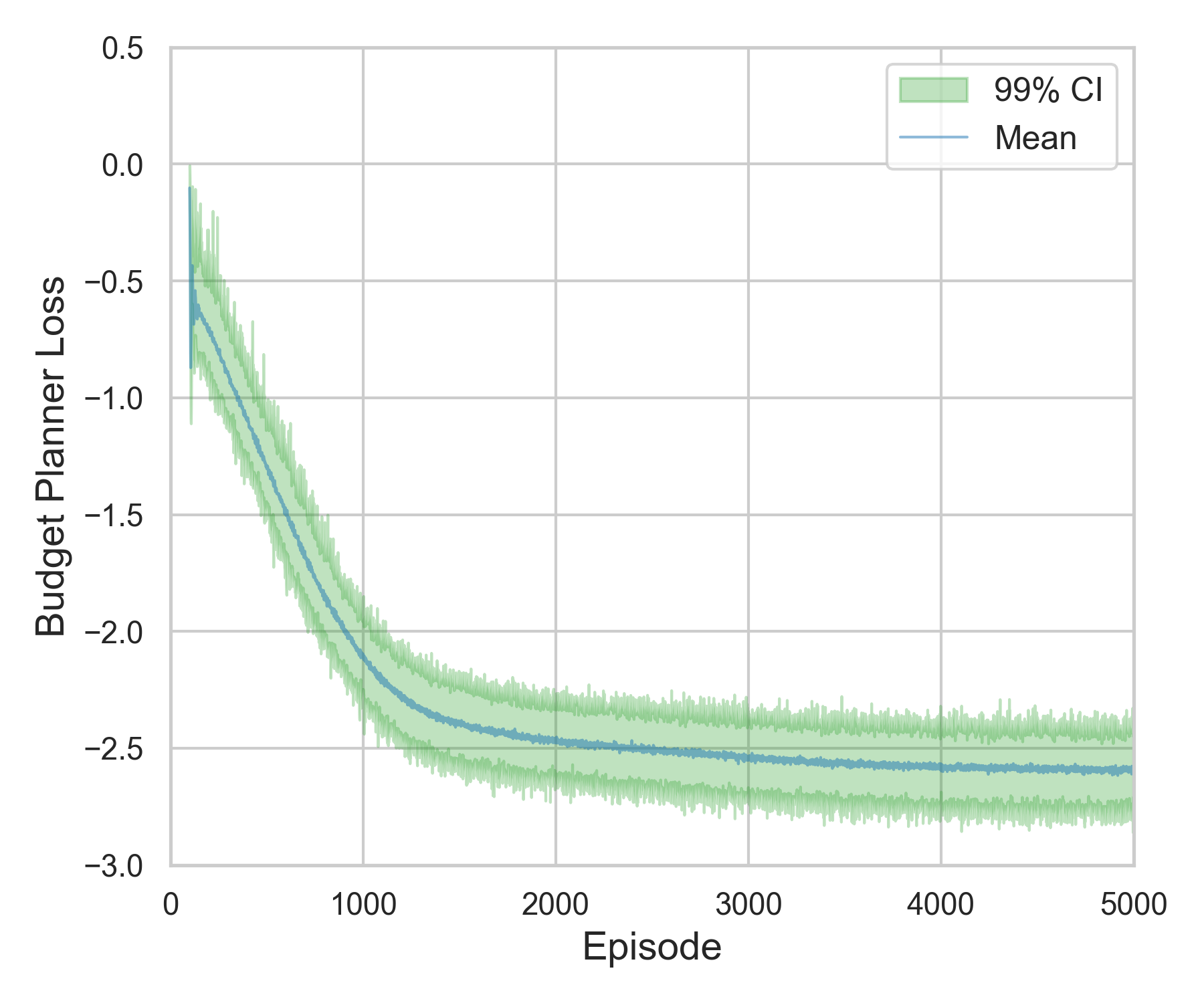}
        \caption{Actor~1 (Budget Planner) Loss}
        \label{fig:actor_loss1}
    \end{subfigure}
    \hfill
    \begin{subfigure}[b]{0.48\textwidth}
        \includegraphics[width=\linewidth]{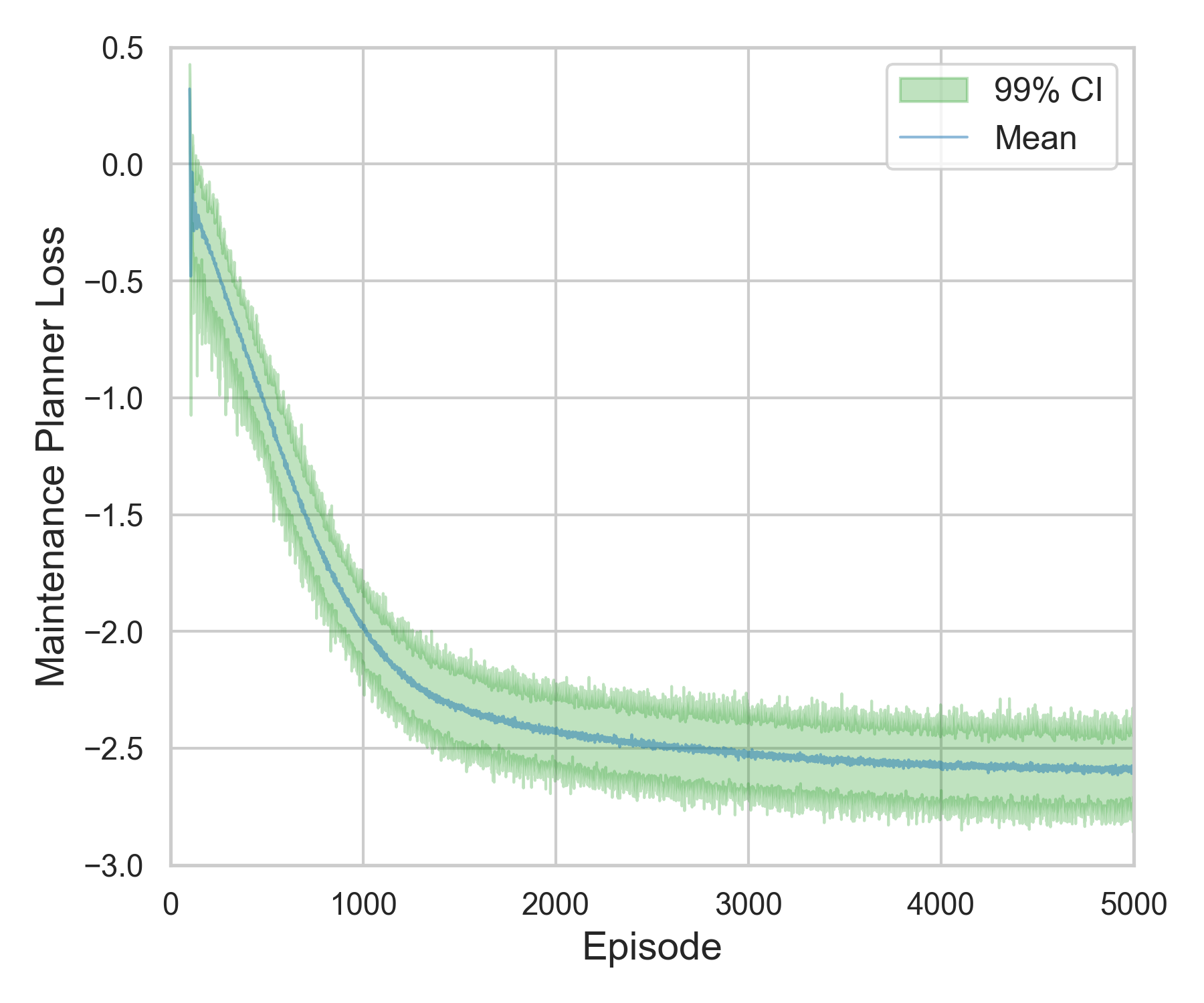}
        \caption{Actor~2 (Maintenance Planner) Loss}
        \label{fig:actor_loss2}
    \end{subfigure}
    \caption{HDRL training dynamics for the 10-sewershed case}
    \label{fig:training_dynamics_2x2}
\end{figure}

For the hierarchical approach, the Budget Planner and Maintenance Planner are trained via a soft-actor–critic framework with separate learning rates: \(8 \times 10^{-4}\) for Actor~1 (budget), \(8 \times 10^{-6}\) for Actor~2 (maintenance), \(1.6 \times 10^{-3}\) for the critic, and \(1.2 \times 10^{-3}\) for the entropy coefficient \(\alpha\). Each network has two hidden layers of size 256, and a total of 5{,}000 episodes are run. The resulting training time is about 391 seconds on average. Figure~\ref{fig:training_dynamics_2x2} illustrates sample training statistics for HDRL in the 10-sewershed setting. Subfigure~(a) shows how the entropy coefficient \(\alpha\) evolves over 5{,}000 episodes, beginning around 0.1 and gradually decreasing toward zero as the policy learns to reduce excessive exploration. Subfigure~(b) depicts the maximum critic loss per mini-batch update on a log scale; a downward trend indicates that the Q-value approximators are becoming more stable and accurate over time. Subfigure~(c) highlights how the Budget Planner's (Actor~1) loss decreases to around \(-2.5\), while subfigure~(d) focuses on the Maintenance Planner's (Actor~2) loss, which follows a similar downward trajectory. Each actor's loss captures the interplay of maximizing Q-values while preserving sufficient policy entropy, leading to improved policies as training proceeds.

\subsubsection{Results Comparison}
\begin{figure}[H]
    \centering
    \includegraphics[width=1.0\columnwidth]{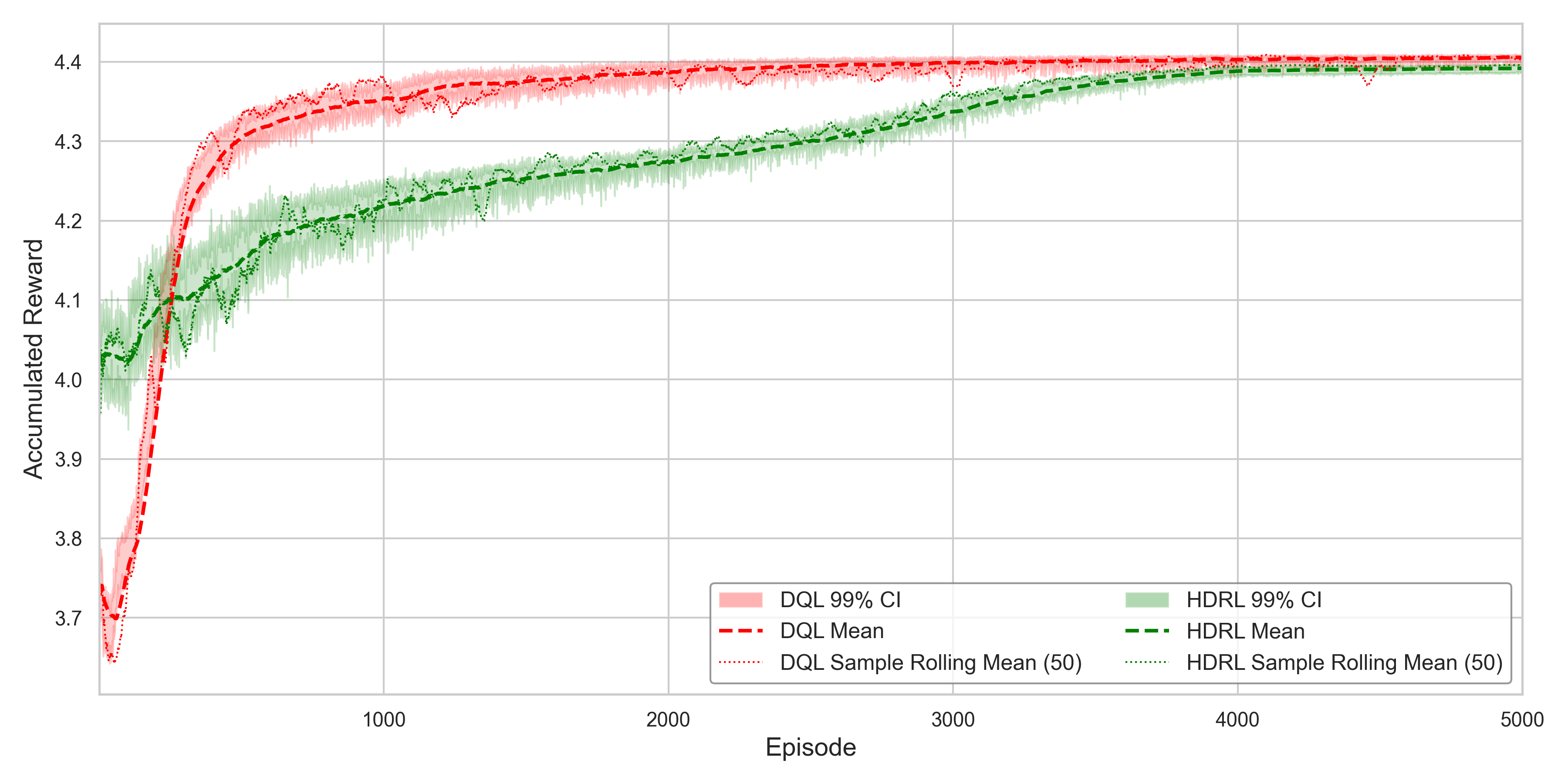}
    \caption{Training Performance (10-Sewershed Network)}
    \label{fig:training_reward_comparison_10c}
\end{figure}

Figure~\ref{fig:training_reward_comparison_10c} shows the training curves for both DQL and HDRL (10 sewersheds, 5{,}000 episodes). The horizontal axis denotes episode count in the training process, and the vertical axis depicts accumulated reward for each episode. DQL begins with faster initial convergence; HDRL stabilizes more uniformly, indicating smoother learning dynamics.

\begin{figure}[H]
    \centering
    \includegraphics[width=0.95\columnwidth]{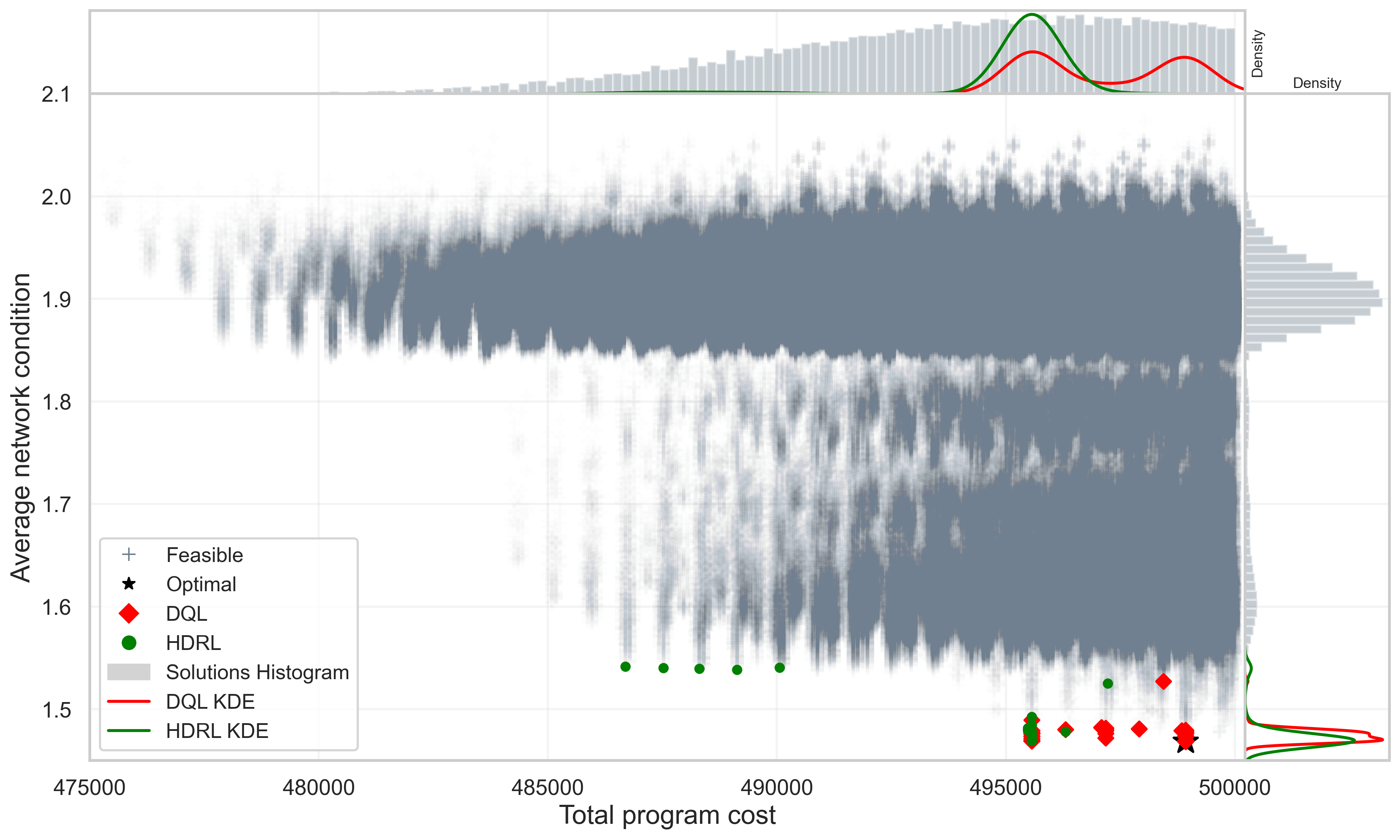}
    \caption{Final cost-condition outcomes of HDRL and DQL for the 10-sewershed network (100 trials each). Marginal KDE curves and histogram illustrate solution distributions along cost (top) and condition (right).}
    \label{fig:jointplot}
\end{figure}

Final policies are assessed after 100 independent training runs for each method.  Figure~\ref{fig:jointplot} illustrates the final solutions (100 runs per method) in the cost-condition space for the 10-sewershed network, where lower condition is preferable. The vast cloud of gray crosses represents all \(2{,}249{,}947\) budget-feasible solutions for a five-year planning horizon, reflecting how complex the feasible landscape can be \citep{Fard2024}. Each cross marks a distinct total program cost (horizontal axis) and the corresponding average sewer network condition (vertical axis). The black star indicates the global optimum, achieving \(\$498{,}925\) total cost and an average condition of \(1.4687\). Red dots denote 100 solutions generated by DQL; green dots show 100 HDRL solutions. Along the top and right edges, marginal histograms and kernel density estimates (KDEs) depict the distribution of costs and conditions for all solutions, revealing how each method clusters within the broader feasible region.

\begin{figure}[H]
    \centering
    \includegraphics[width=0.95\columnwidth]{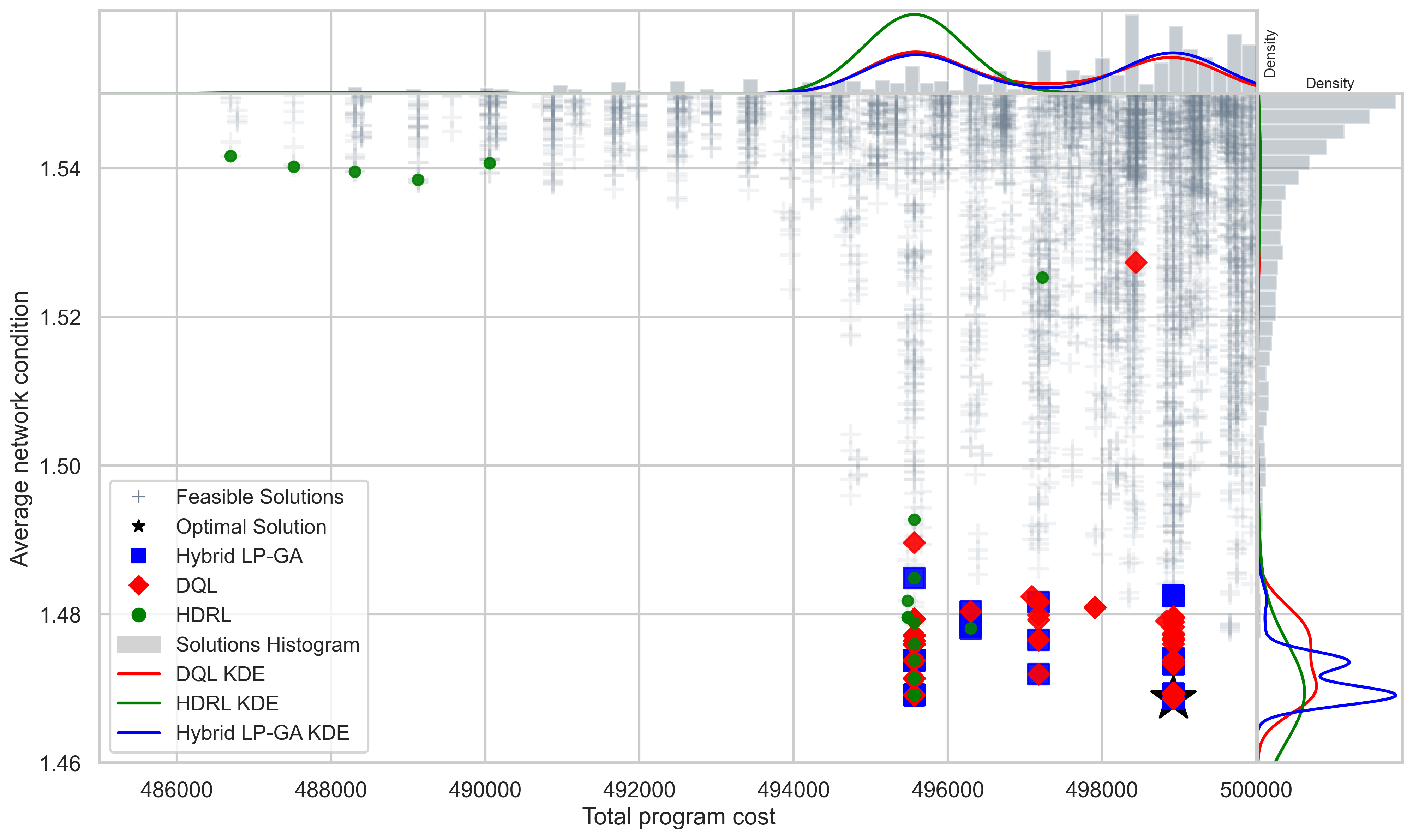}
    \caption{Zoomed view of the near-optimal region comparing HDRL, DQL, and Hybrid LP-GA in cost-condition space (10 sewersheds). Marginal KDEs along the axes highlight the density of solutions for each method.}
    \label{fig:jointplot-mag}
\end{figure}

Figure~\ref{fig:jointplot-mag} zooms in on the near-optimal region of the same cost-condition plane and includes additional 100 results generated by a Hybrid LP-GA method \citep{Fard2024}. The Hybrid LP-GA algorithm couples local linear programs with a global evolutionary search, thereby producing feasible interventions with high solution quality. among these 100 solutions for each method DQL 9 times found the global optimum solution and Hybrid LP-GA found the global optimum 11 times. However the HDRL didnt find the global optimal in 100 attemps, however in 86 atempts reach to a unique solution (1.4691 , \(\$495{,}564\)). The zoomed view shows that HDRL (green) tends to find the near-optimal solutions with lower cost compared to  DQL (red) solutions and Hybrid LP-GA (blue). Mean condition values over the 100 trained models are 1.4723 for HDRL, 1.4744 for DQL, and 1.4716 for LP–GA. This figure underscores how HDRL compares favorably with both conventional RL (DQL) and advanced metaheuristic methods, maintaining strong performance close to the global optimum.  As can be seen in Figure~\ref{fig:jointplot} the HDRL method found a few solutions with lower cost found on the Pareto front of cost-condition space.

\subsection{Scaling to 15 and 20 Sewersheds}

Two larger case studies are examined to test scalability. Increasing from 10 to 15 or 20 sewersheds adds complexity in both the state space (due to additional components) and the action space (due to more potential decisions).

\begin{figure}[H]
    \centering
    \includegraphics[width=1.0\columnwidth]{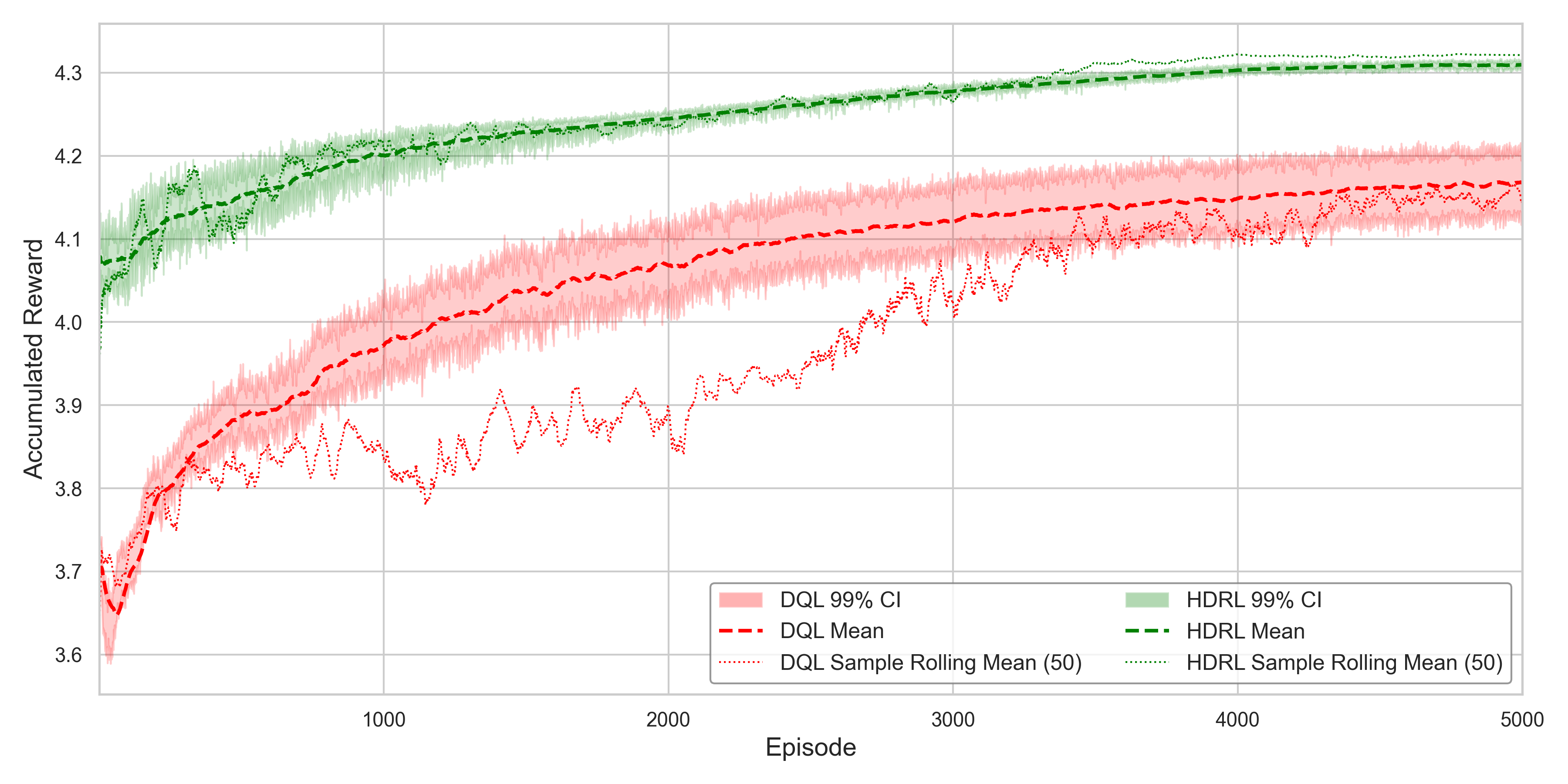}
    \caption{Training Performance (15-Sewershed Network)}
    \label{fig:training_reward_comparison_15c}
\end{figure}

For 15 sewersheds, there are \(2^{15} = 32{,}768\) nominal flush/no-flush combinations at each decision step, but only 364 actions satisfy the annual budget constraint. This means the DQL agent can constrain its Q-network output to 364 instead of 32{,}768. Figure~\ref{fig:training_reward_comparison_15c} depicts DQL (red) and HDRL (green) performance over 5{,}000 episodes. DQL gradually improves but exhibits higher variance, whereas HDRL maintains more stable returns. In HDRL, the Budget Planner’s actor output dimension stays at 1, and the Maintenance Planner’s actor output size scales linearly with the number of sewersheds. By contrast, DQL’s output layer size grows exponentially with the number of sewersheds, indicating that HDRL’s split between budget decisions and maintenance actions is more scalable.

\begin{figure}[H]
    \centering
    \includegraphics[width=1.0\columnwidth]{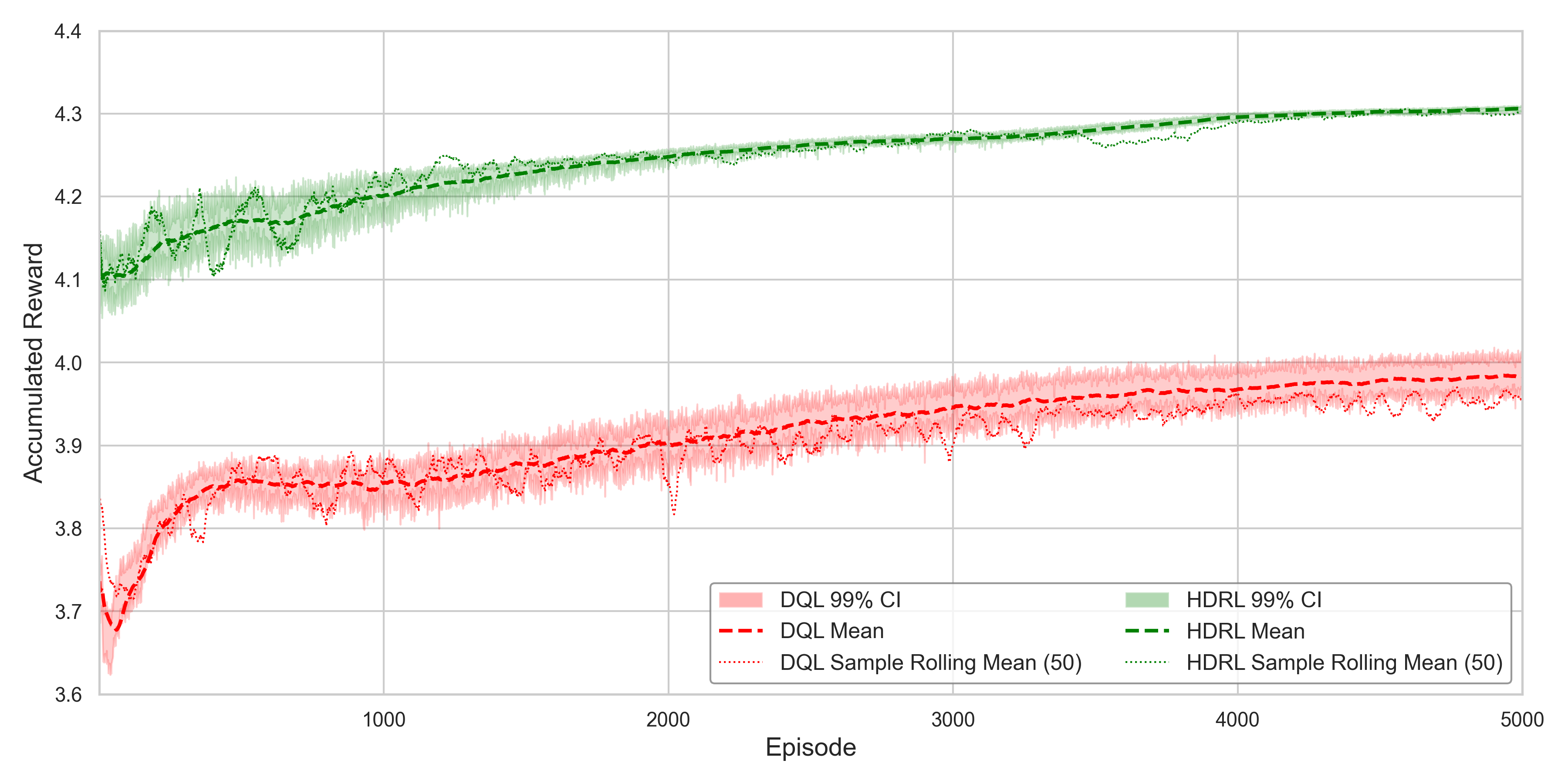}
    \caption{Training Performance (20-Sewershed Network)}
    \label{fig:training_reward_comparison_20c}
\end{figure}

When expanded to 20 sewersheds, the baseline combinatorial action space has \(2^{20} = 1{,}048{,}576\) possibilities per decision, cut down to 7{,}448 by the budget filter. Figures~\ref{fig:training_reward_comparison_20c} display DQL and HDRL training curves. DQL experiences pronounced fluctuations as it contends with a large discrete action set. HDRL still converges more smoothly and achieves higher returns. Again, the hierarchical approach keeps Actor~1’s output fixed at a single budget fraction and grows Actor~2’s output only linearly with the network size. In contrast, DQL faces a potentially exponential expansion in its final Q-network layer, which burdens training.

HDRL assigns one actor to produce a yearly budget fraction and another to allocate that budget across the individual sewersheds. This decomposition yields a constant-size output for the Budget Planner’s actor and critic, and a linearly expanding output for the Maintenance Planner. By contrast, DQL must handle an exponentially growing action space at the final Q-network layer as more sewersheds are added. The hierarchical design thereby reduces computational overhead and fosters more stable training, as evidenced by the smoother reward curves and higher final returns under HDRL. Although hyperparameters and domain features can influence results, these experiments show that HDRL consistently retains its advantage as the problem size grows from 10 to 15 and eventually to 20 sewersheds.

\subsection{Discussion of Scalability}

Table~\ref{tab:scalability_summary} summarises how the two methods respond as the network size grows from 10 to 20 sewersheds.  Three patterns stand out.  First, the dimensionality of DQL’s final layer rises exponentially: 20 outputs for the smallest instance, 364 at 15 sewersheds, and 7 448 at 20.  HDRL, by contrast, always predicts a single budget fraction and one priority logit per asset, so its output size grows only linearly (\(1+n\)).  Second, this difference in network width translates into markedly different training costs.  While HDRL is slower on the 10-sewershed case because it must update two actors, two critics, and solve an embedded linear programme, its runtime increases only modestly thereafter (from 264 s to 585 s for 5 000 episodes).  DQL, however, sees its runtime almost quintuple over the same range as it struggles to back-propagate through thousands of Q-values.  Finally, solution quality diverges as the problem scales.  On the smallest instance the average objective values of the two methods are nearly identical, but with 15 sewersheds HDRL is already about eight per cent better, and on 20 sewersheds the gap widens to roughly fifteen per cent.  In short, hierarchical decomposition curbs the combinatorial explosion that undermines DQL, allowing HDRL to maintain stable learning and superior outcomes even as the state and action spaces expand dramatically.

\begin{table}[H]
    \centering
    \caption{Comparison of DQL and HDRL Performance Across 10-, 15-, and 20-Sewershed Networks}
    \label{tab:scalability_summary}
    \small
    \renewcommand{\arraystretch}{1.1}
    \begin{tabular}{lcccc}
        \hline\hline
        \textbf{Case} & \textbf{Method} & \textbf{Output Layer Size} & \textbf{Runtime (s)} & \textbf{Avg.\ Obj.\ Value} \\
        \hline
        10 sewersheds
            & DQL  
              & 20 
              & 118 $\pm$ 4  
              & 1.4744  \\ 
            & HDRL 
              & 1 $+$ 10\(^\dagger\) 
              & 264 $\pm$ 6
              & 1.4723 \\[1em]
        15 sewersheds
            & DQL  
              & 364  
              & 176 $\pm$ 5
              & 1.6629 \\ 
            & HDRL 
              & 1 $+$ 15 
              & 366 $\pm$ 6
              & 1.5368 \\[1em]
        20 sewersheds
            & DQL  
              & 7{,}448 
              & 570 $\pm$ 5
              & 1.8128 \\
            & HDRL 
              & 1 $+$ 20 
              & 585 $\pm$ 7
              & 1.5375 \\
        \hline\hline
        \multicolumn{5}{l}{\footnotesize 
    \parbox[t]{0.95\linewidth}{%
        \(^\dagger\)~HDRL uses one-dimensional output for the Budget Planner (Actor~1) plus $n$-dimensional output for the Maintenance Planner (Actor~2).
    }%
}\\
    \end{tabular}
    \normalsize
\end{table}

\section{Conclusion}

This paper presents an HDRL framework designed to tackle multi-year infrastructure maintenance planning under stringent budget constraints. The method decomposes the decision space into two levels: a Budget Planner that selects an annual budget within feasible bounds and a Maintenance Planner that allocates that budget across individual assets. This hierarchical design avoids the exponential action blow-up commonly encountered by monolithic RL methods with combinatorial action, making HDRL more tractable for larger networks. A local linear programming step further ensures that annual cost remains within the chosen budget. 

Comparisons with a Deep Q-Learning (DQL) baseline, across 10-, 15-, and 20-sewershed case studies, show that HDRL yields smoother training performance, converges more reliably, and consistently discovers near-optimal or high-performance maintenance policies. In the 10-sewershed instance, the HDRL solutions cluster tightly near global optima solution obtained by constraint-programming, and in the 15 and 20-sewershed extension, HDRL continues to learn stable policies while DQL’s performance deteriorates due to the exponentially larger action space. Empirical results confirm that dividing budget allocation from local maintenance selection significantly reduces complexity for each actor network and maintains stable learning even as network size grows. This capacity to preserve scalable action spaces with explicit budget enforcement addresses a longstanding gap in reinforcement learning applications to real-world asset management.

Future work may extend HDRL by incorporating partial observability of asset condition (e.g., uncertain inspection data), modeling dynamic climate or demand scenarios, and exploring different forms of hierarchical decomposition (multi-level budgets or region-based planners). Additional refinements, such as multi-agent collaboration, dynamic reward shaping, or specialized optimization routines, might further enhance performance in extremely large-scale systems with tens of thousands of assets. Nonetheless, the presented approach provides a sound foundation for bridging the gap between theoretical DRL advances and the practical demands of infrastructure maintenance planning under tight financial limits.

\section*{Acknowledgements}
The authors gratefully acknowledge the partial financial support provided by the Natural Sciences and Engineering Research Council of Canada (NSERC) Discovery Grant (RGPIN-2022-04591).

\section*{Reproducibility.}
All source code and the processed sewer-network datasets used in our experiments are publicly available in a GitHub repository for review at \url{https://github.com/amirkfard/Hierarchical-Deep-Reinforcement-Learning}.

\bibliographystyle{apacite}   
\bibliography{bibliography}

\end{document}